\documentclass[journal]{IEEEtran}

\ifCLASSINFOpdf
  \usepackage[pdftex]{graphicx}

\else

\fi

\usepackage{amsmath}

\usepackage{background}
\backgroundsetup{
  position=current page.east,
  angle=-90,
  nodeanchor=east,
  vshift=-5mm,
  hshift = 92mm,
  opacity=1,
  scale=1.35,
  contents=Non-peer reviewed. This work has been submitted to IEEE Transactions on Geoscience and Remote Sensing for possible publication.
}

\begin{document}

This work has been submitted to the IEEE for possible publication. Copyright may be transferred without notice, after which this version may no longer be accessible.

\pagebreak

\title{Improving the Thermal Infrared Monitoring of Volcanoes: A Deep Learning Approach for Intermittent Image Series}

\author{Jeremy~Diaz,
        Guido~Cervone,
        and~Christelle~Wauthier%
\thanks{J. Diaz is currently with the United States Geological Survey Water Mission Area. At the time of this work, he was with the Department of Geography and the Institute of Computational and Data Sciences, The Pennsylvania State University, University Park,
PA, 16801 USA (e-mail: jad6655@psu.edu)}%
\thanks{G. Cervone is with the Department
of Geography and the Institute of Computational and Data Sciences, The Pennsylvania State University, University Park,
PA, 16801 USA}%
\thanks{C. Wauthier is with the Department
of Geosciences and the Institute of Computational and Data Sciences, The Pennsylvania State University, University Park,
PA, 16801 USA}%
\thanks{Manuscript received April 19, 2005; revised August 26, 2015.}}

\markboth{Journal of \LaTeX\ Class Files,~Vol.~14, No.~8, August~2015}%
{Shell \MakeLowercase{\textit{et al.}}: Bare Demo of IEEEtran.cls for IEEE Journals}

\maketitle

\begin{abstract}
    Active volcanoes are globally distributed and pose societal risks at multiple geographic scales, ranging from local hazards to regional/international disruptions. Many volcanoes do not have continuous ground monitoring networks; meaning that satellite observations provide the only record of volcanic behavior and unrest. Among these remote sensing observations, thermal imagery is inspected daily by volcanic observatories for examining the early signs, onset, and evolution of eruptive activity. However, thermal scenes are often obstructed by clouds, meaning that forecasts must be made off image sequences whose scenes are only usable intermittently through time. Here, we explore forecasting this thermal data stream from a deep learning perspective using existing architectures that model sequences with varying spatiotemporal considerations. Additionally, we propose and evaluate new architectures that explicitly model intermittent image sequences. Using ASTER Kinetic Surface Temperature data for $9$ volcanoes between $1999$ and $2020$, we found that a proposed architecture (ConvLSTM + Time-LSTM + U-Net) forecasts volcanic temperature imagery with the lowest RMSE ($4.164^{\circ}$C, other methods: $4.217-5.291^{\circ}$C). Additionally, we examined performance on multiple time series derived from the thermal imagery and the effect of training with data from singular volcanoes. Ultimately, we found that models with the lowest RMSE on forecasting imagery did not possess the lowest RMSE on recreating time series derived from that imagery and that training with individual volcanoes generally worsened performance relative to a multi-volcano data set. This work highlights the potential of data-driven deep learning models for volcanic unrest forecasting while revealing the need for carefully constructed optimization targets.
\end{abstract}


\IEEEpeerreviewmaketitle

\section{Introduction}

\IEEEPARstart{F}{orecasting} volcanic unrest has been a major scientific goal  for at least 100 years \cite{Poland2020}. This is important because over 800 million people live within 100 kilometers of a Holocene volcano \cite{Brown2015a}. Beyond direct volcanic hazards, volcanoes can perturb agriculture, infrastructure, the environment, as well as (inter)national business and travel \cite{Brown2015} during the course of an eruption or during inter-eruptive periods. Short-term (e.g., days to months) volcanic unrest forecasts are made by identifying patterns in monitoring observational time series (e.g. seismic, deformation, thermal, emissions). Improving these forecasts has been identified as a grand challenge within the field of volcanology \cite{NASEM2017}. Ultimately, this task is very challenging because there are various methods, expertise, and/or instrumentation to monitor volcanoes on the ground, as well as remotely from the air and space (see \cite{NASEM2017} section 1.4 for a light overview, see \cite{Reath2019} for a more detailed remote sensing perspective). As a result, reliable and timely monitoring and, thus, the ability to forecast unrest accurately is costly and absent from many active volcanoes \cite{Brown2015}.

Short-term forecasting of volcanic unrest is conducted through a range of domain and statistical approaches, both quantitative and qualitative. At the highest level, these forecasts are generated by expert committees at observatories \cite{Aspinall2010} and described through event trees which communicate all possible outcomes with the associated probabilities at each stage of the outcome \cite{Newhall2002}. Likewise, the very few examples of quantitative methods which directly output broadly prescriptive forecasts (i.e., many metrics or probabilities) rely upon event trees as input data \cite{Marzocchi2008, Junek2012}. These event tree approaches are very promising but also very demanding, requiring a plethora of expert knowledge, monitoring streams, and volcano-specific knowledge. On the other hand, quantitative methods are often more limited in scope, instead being applied to individual time series of monitoring data to aid in the identification of early warnings of change (e.g., values above background, changes in pattern), and this exercise is often considered forecasting, see \cite{Swanson1985, Poland2020a}. Examples of this approach that are heavily influenced by domain knowledge include modeling seismic signals from a material failure perspective \cite{DelaCruz-Reyna2001, Lavallee2008} and monitoring gas ratios \cite{Aiuppa2007} and lava discharge rates \cite{Ripepe2017} from an equilibrium/disequilibrium perspective. On the other hand, more purely quantitative approaches include the use of simple thresholds \cite{Poland2020}, examination of current events among their broader distributions \cite{Marzocchi2012}, and various modeling projects that have made use of Poisson point processes \cite{DelaCruz-Reyna1991, Bebbington2013}, variograms \cite{Jaquet2006, Carniel2008}, Markov chain Monte Carlo methods \cite{Segall2013}, and ARIMA models \cite{Ho2008, Carla2016}.

With global volcanic monitoring in mind, there are many difficulties and opportunities with this existing toolkit of forecasting options. Particularly, individual forecasting efforts generally differ in the variety and historical reach of their available data. Committee-based and event tree approaches are particularly ambitious, but their reliance on many experts makes them inaccessible due to global volcanic monitoring already being financially restrictive. Likewise, quantitative methods that employ domain knowledge often make use of specialized ground equipment (e.g., seismometers, infrared spectrometers) that is lacking at many active volcanoes. Thus, financially accessible solutions which are globally homogeneous are highly appealing. Because of this, satellite remote sensing has been thoroughly studied \cite{Furtney2018} and widely adopted to both establish and supplement monitoring efforts \cite{Reath2019a, Ramsey2016, Patrik2016, Wright2002}. Currently, multiple online tools exist to aid global volcanic monitoring via remote sensing; examples of these include MODVOLC \cite{Wright2002,Wright2016}, MIROVA (Middle InfraRed Observation of Volcanic Activity) \cite{Coppola2016}, and MOUNTS (Monitoring Unrest From Space) \cite{Valade2019}. Thermal monitoring has seen particular operational success due to the relative ease of access, processing, and interpretation. This imagery is often provided for viewing while accompanied by time series of higher-level information, such as the maximum temperature above the background (i.e., ``excess temperature’’) \cite{Wright2002}, the number of hotspots \cite{Valade2019}, the maximum distance of a hotspot from the summit \cite{Coppola2016}, and/or measures of radiance \cite{Coppola2016}.

Thermal remote sensing imagery is therefore a promising data stream for global volcanic monitoring and forecasting. However, for a predictive image task, the routine statistical models mentioned earlier have multiple weaknesses: notably, they make rigid assumptions about the functional relationships of data and the distributions underlying those data, do not necessarily consider spatially correlated data, and, perhaps most importantly, are relatively simple\textemdash exploring a parameter space much smaller than current computational technology allows. Meanwhile, since 2012 \cite{Krizhevsky2012}, neural networks have seen an explosive growth in popularity (including within the field of remote sensing \cite{Ma2019c}) due to their incredible performance in making image-based predictions by utilizing millions of optimized parameters and lots of data. Likewise, machine/deep learning has been applied in the field of volcanology for various detection and classification tasks (particularly with seismic signals and Interferometric Synthetic Aperture - InSAR - geodetic data) \cite{IbsvonSeht2008, Curilem2009, Piscini2014, Anantrasirichai2018, Valade2019, Sun2020}, but the task of forecasting remains poorly explored  \cite{Ma2019c,Poland2020}. However, in the existing examples of short-term volcanic forecasting with machine learning \cite{Brancato2016, Nomura2020, Dempsey2020}, neural networks have more commonly excelled \cite{Brancato2016, Nomura2020} and among neural networks, more sophisticated architectures have proven beneficial \cite{Nomura2020}. Lastly, while it is undesirable that neural networks are infamous for having uninterpretable mappings of input data to output targets, they are additionally favorable for this cloud-susceptible task because they have a documented history for performing well with noisy and missing data \cite{Openshaw2000}.

In summary, improving short-term volcanic forecasts are valuable for society worldwide. There are many ways to produce these forecasts, but most cannot be implemented globally due to the costs of experts and specialized ground equipment. There are, however, many globally accessible remote sensing data streams; particularly, satellite thermal imagery has been widely adopted. Forecasting remotely sensed imagery\textemdash especially imagery that can be inhibited by atmospheric conditions\textemdash is not well explored. Here, we draw on existing and proposed models (with an emphasis on neural networks) to explore and evaluate the task of forecasting volcanic temperature imagery. Specific experiments include:

\begin{itemize}
    \item Which model (among naive methods, autoregression, and numerous neural networks) can best forecast volcanic AST\_08 image sequences?
    \item How do derived time series from these models compare to real observations?
    \item How does performance on individual volcanoes change when the training data set is limited to individual volcanoes instead of all 9 in the data set?
\end{itemize}

\section{Data and Methods}

\subsection{Data Set}

Nine active volcanoes were targeted in this project: Mount Erebus (Antarctica), Erta Ale (Ethiopia), Mount Etna (Italy), Kīlauea (Hawaii), Masaya (Nicaragua), Nyamulagira (Democratic Republic of Congo), Nyiragongo (Democratic Republic of Congo), Pacaya (Guatemala), and the distant Pu'u 'Ō'ō cone of Kīlauea (Hawai’i, US). These volcanoes were chosen for their active nature and/or because they have displayed lava lakes (a superficial and persistent thermal phenomenon) for at least some percentage of the data record. Among these, the Hawaiian volcanoes and Mount Etna are among the most studied and monitored in the world; Erebus, Masaya, and Pacaya have some degree of ground monitoring; Nyamulagira and Nyiragongo have decommissioned ground monitoring efforts due to civil unrest in the region; and, Erta Ale does not have regular ground monitoring \cite{Brown2015a}.

Among thermal monitoring options, nighttime imagery is preferred in the field of volcanology because it is generally less cloudy and decreases the effects of topography and solar heating, ``allowing better delineation of low-temperature geothermal anomalies'' \cite[p. 18]{Pieri2004}. In general, many platforms do not make nighttime thermal imagery easily accessible (e.g., Landsat has special request targets \cite{Arvidson2001}); however, the $>20$-year NASA Terra platform which hosts the MODIS and ASTER instruments are an exception to this. MODIS is very popular because it provides a very high temporal sampling rate (four images per day \cite{Coppola2020}) which can aid in the rapid detection of changes \cite{Wright2016}. ASTER, on the other hand, provides a 10-fold increase in spatial resolution which allows the less frequent detection of subtler changes months in advance \cite{Pieri2005, Reath2016, Reath2019}. Nighttime thermal data from ASTER (L1T) 
is available for bulk download in the form of 5 bands of radiance values (8-12$\mu$m) from the online DAAC2Disk Utility, courtesy of the NASA Land Processes Distributed Active Archive Center (LP DAAC), USGS Earth Resources Observation and Science (EROS) Center, Sioux Falls, South Dakota, https://lpdaac.usgs.gov/tools/daac2diskscripts/. The level-2 Surface Kinetic Temperature product (AST\_08) \cite{NASA2001} which is more thoroughly corrected for the atmosphere and with radiance converted to temperature is available for order from the NASA LP DAAC via NASA's Earthdata Search (https://search.earthdata.nasa.gov/search?q=ast\_08).
	
First, all L1T data for each volcano between the dates December 18, 1999, and May 1, 2020, were downloaded. Initial inspection revealed a poor correlation between viable scenes and cloud cover metadata, so all scenes were manually inspected for viability in the data set. Upon manual inspection, each scene was denoted viable, uncertain, or nonviable (\textbf{Figure 1)}. Due to the large number of nonviable scenes, uncertain scenes were ultimately included in our data set; these were scenes with partial cloud cover or scenes with large extents of cloud cover, through which, thermal features were still observable. The final set of viable and uncertain scenes were then ordered as level-2 AST\_08 data for the remainder of the project. The distribution of time between these scenes is presented in \textbf{Figure 2}.

\begin{figure}[t]
\centering
\includegraphics[width=\linewidth]{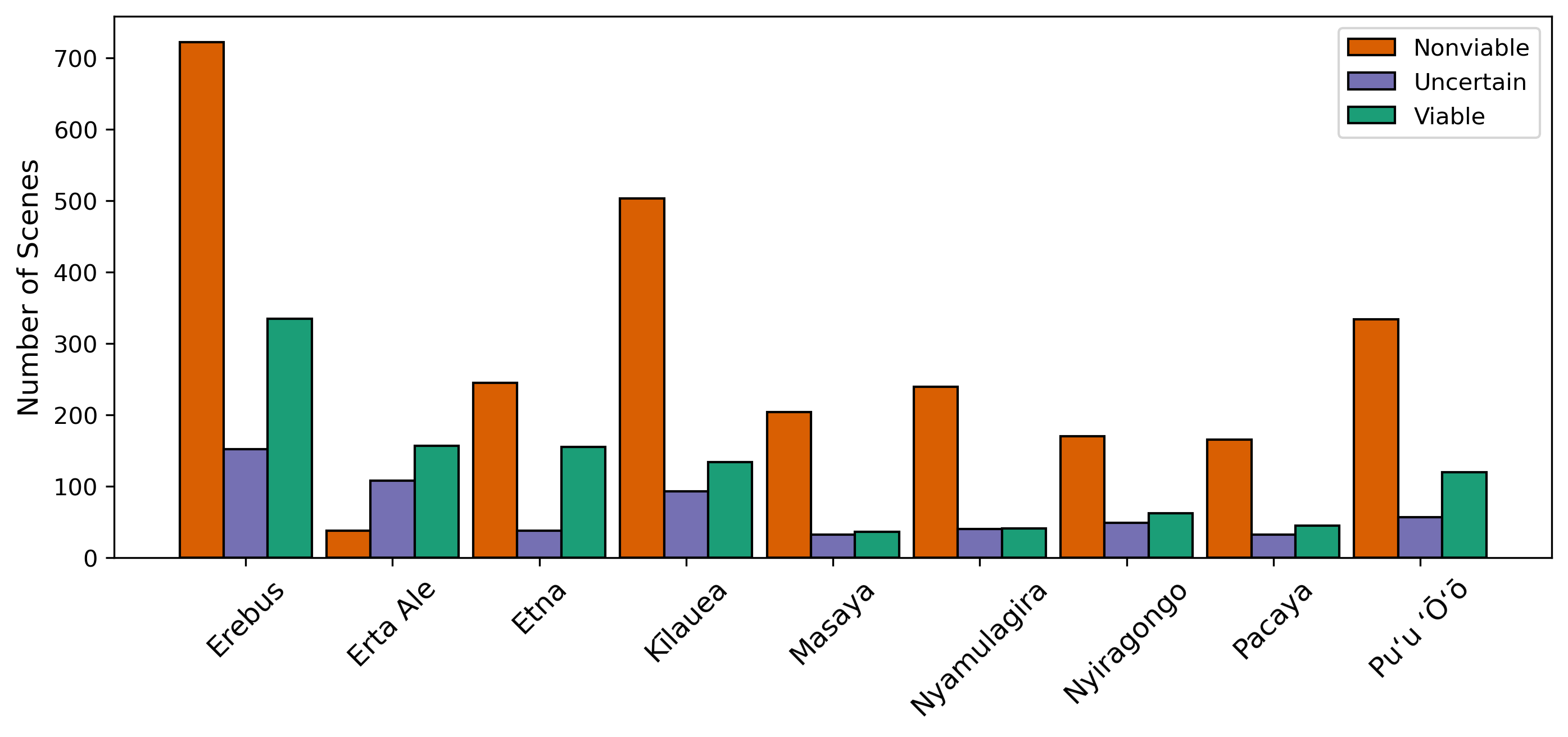}
\caption{Manual scene inspection results by volcano. Scenes are from the beginning of ASTER record to May 1, 2020.}
\end{figure}

\begin{figure}[t]
\centering
 \includegraphics[width=\linewidth]{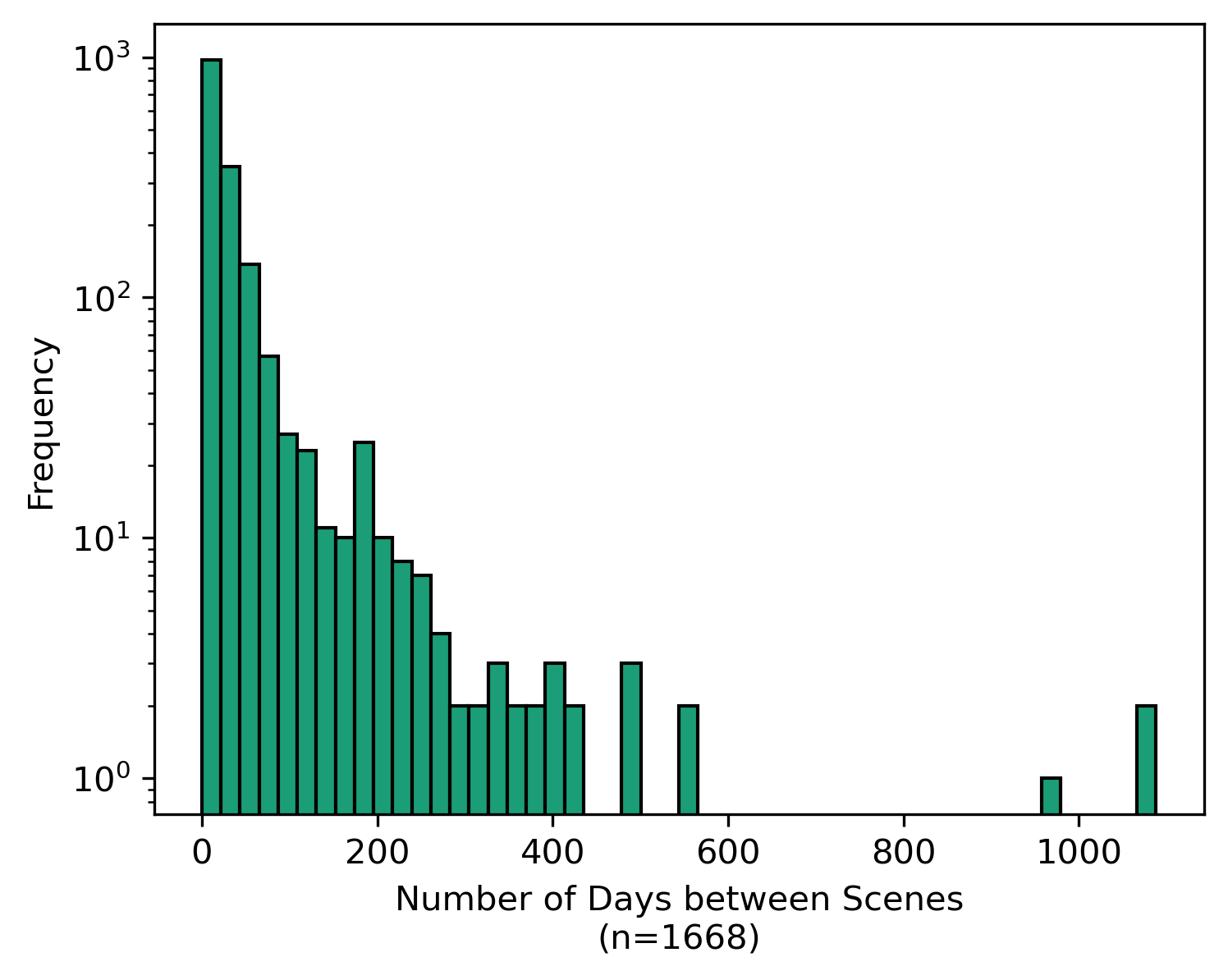}
\caption{Distribution of the amount of time between scenes in this data set. Mean = $37.138$. Standard deviation = $73.850$.}
\end{figure}
	
\subsection{Data Processing}

The AST\_08 data contained missing values representing areas outside the sensor’s field of view as well as pixels amid very volcanically active areas (see \textbf{Figure 3}, top row) – with sources referencing the latter as recovery pixels \cite{Rose2009}. In the level-2 data, the latter appears as counter-intuitive dark spots within a bright hotspot, while the level-1 data does not contain these missing pixel observations. Since locations outside the field of view will have valid observations in at most two cardinal directions, pixels with valid values in at least three cardinal directions were filled with nearest-neighbor interpolation; qualitatively, this better resembled level-1 data and did not assign improbable "cold" pixels islands in the middle of large lava lakes/flows.

To narrow the focus of the image forecasting task, a $96\times96$ pixel subset of the scene was used in the data set; with a 90-meter spatial resolution, this translates to an $8.64\times8.64$ kilometer subset. This precise size is somewhat arbitrary, but, qualitatively, it was large enough to observe significant portions of large lava flows while not being too large to neglect small lava lakes. The subset was centered on the latitude and longitude of the volcano summit. Converting the scenes’ coordinate reference systems to latitude and longitude for summit locating and subsetting introduced geometric distortions that were visually apparent for Mount Erebus (located in Antarctica); however, these relics were not problematic for thermal interpretation. 

In volcanology studies, temperature measurements are often examined relative to the environmental background (i.e., the excessive temperature of the volcanic feature relative to the surrounding environment); as such, the AST\_08 scenes were processed this way (see \textbf{Figure 3} for steps up to this point). Volcano studies often manually identify a cloud-free, representative part of the environment to determine the environmental background temperature. To automate this process in a robust way, the following process was used to compute the average environmental background temperature: first, a $10\times10$ pixel subset was extracted from each corner of the reduced image; these corners were then considered appropriate if less than 10\% of their pixel values were outside the sensor’s field of view (i.e., missing values). The mean was then calculated for each subset, and the mean of those means was used as the environmental background temperature. Mount Erebus’s location along the orbital path led to unique viewing geometries where this method would fail to identify any valid corners; in those cases, the $10\times10$ pixel subsets were rotated around the perimeter of the image to identify suitable areas (see \textbf{Figure 4}). Once the environmental background temperature was computed, it was subtracted from every pixel value. Notably, in addition to being consistent with volcanology studies, this removes the task of forecasting the seasonality of the background environment.

\begin{figure*}[t]
\centering
\includegraphics[width=\linewidth]{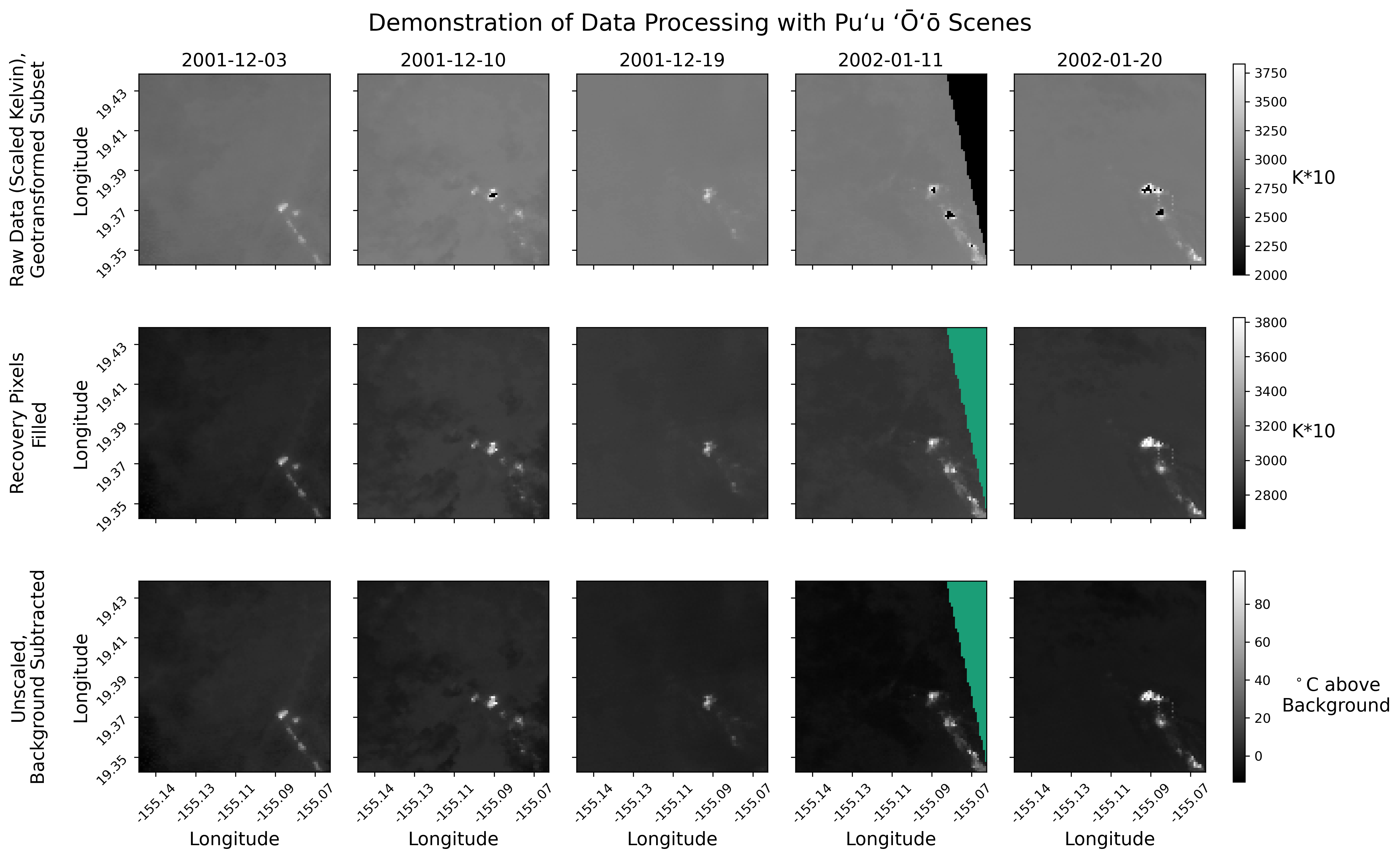}
\caption{Visualization of three steps of the data processing for a five-scene sequence of Pu'u 'Ō'ō. Each row is a stage of data processing, and each column is the same scene. The green shade denotes formally acknowledged missing values.}
\end{figure*}

\begin{figure}[t]
\centering
\includegraphics[width=\linewidth]{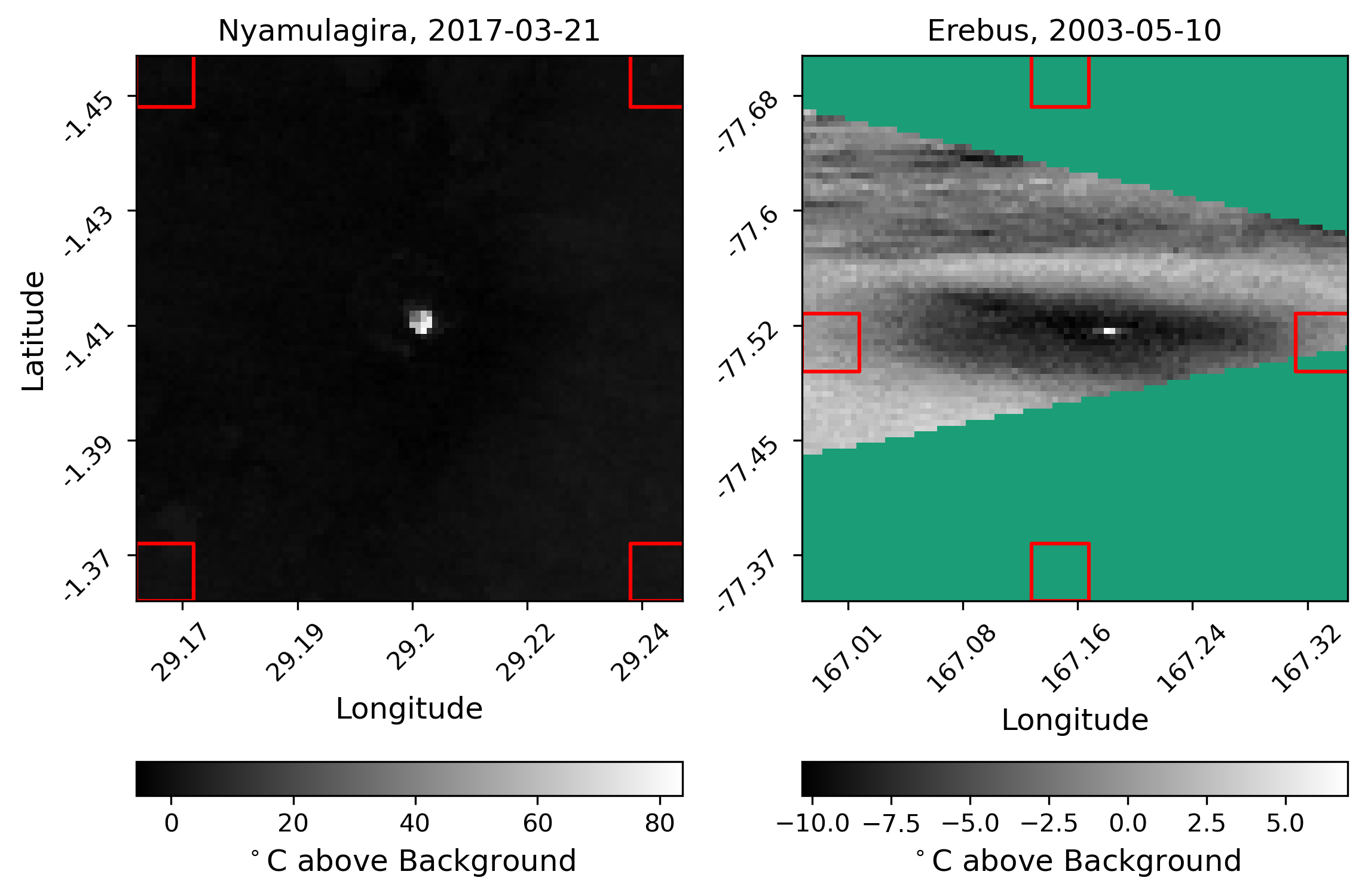}
\caption{The two ways that background areas were defined to calculate the average background temperature of the scene. Red boxes denote the background area, and the green shade denotes formally acknowledged missing values.}
\end{figure}

After the above, there remained missing values from the $96\times96$ scene subsets that represented areas outside the sensor’s field of view. These pixels were then filled with the values from the previous scene and the date of those previous scenes was noted. When this occurred for the first scene acquired for a volcano (i.e., no previous scenes), the missing values were filled with nearest-neighbor interpolation (\textbf{Figure 5}). Lastly, prior to training the data was scaled between 0 and 1 (i.e., $x_{new} = (x_{original} - x_{min}) / (x_{max} - x_{min})$) to facilitate training convergence. 

\begin{figure*}[t]
\centering
\includegraphics[width=0.45\linewidth]{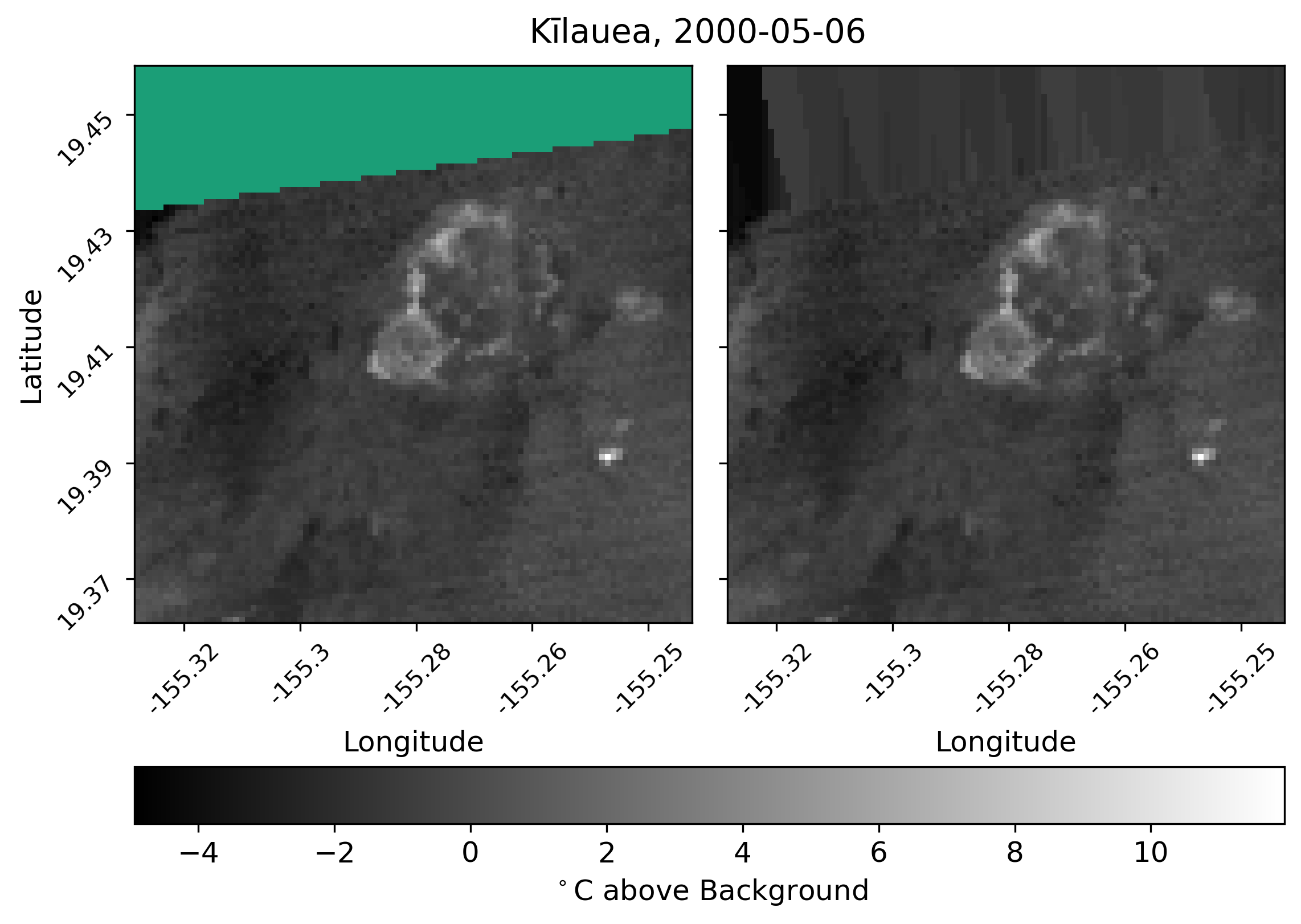}
\includegraphics[width=0.45\linewidth]{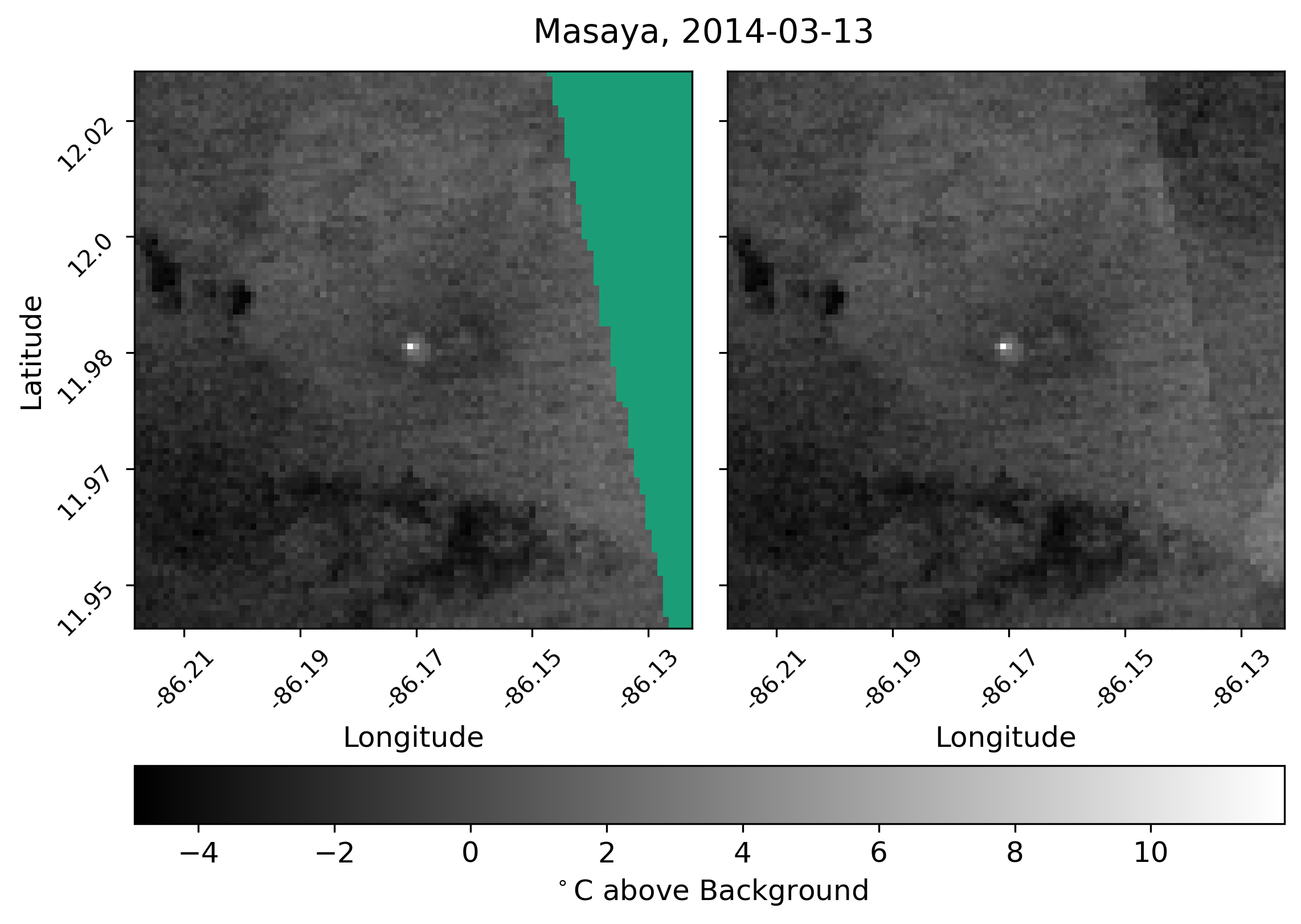}
\caption{The two methods for filling in missing values. Left: when no scenes exist prior to the scene of interest, missing values were filled with nearest neighbor interpolation. Right: when scenes exist prior to the scene of interest, missing values were filled with prior observations while tracking the difference in time.}
\end{figure*}

\subsection{Forecasting Configuration}

The forecasting task here uses the previous $n$ scenes to forecast the next scene. For most models, this is done by treating each pixel as an independent observation, but for convolution-based neural networks, this is done at the image-level. The exact choice of $n$ is not straightforward, but realistically, it must be large enough to make use of past information and small enough to not magnify computational and storage burdens. A popular method in volcanology studies is to examine the maximum excess temperature through time, so these time series were extracted from the L1T data (prior to AST\_08 retrieval) and their autocorrelation structure was examined for each volcano. The average of the largest significant autocorrelation lag was $6.43$, so $n$ was set to $6$.

In the case of volcano-specific training sets (see Results and Discussion, section C), the largest significant autocorrelation lag for each volcano was used with a minimum accepted value of $3$ and a maximum accepted value of $10$ previous scenes.

\subsection{Models}

\subsubsection{Baseline predictive methods}

ARIMA models are common time series models (see \cite{Cryer&Chan2008} for a comprehensive coverage) with published applications in volcanic forecasting \cite{Ho2008, Carla2016}. The ARIMA acronym stands for ``autoregressive integrated moving average'' because it combines the simpler autoregressive (AR) model that predicts $x_{t}$ based on the linear combination of its previous $\{x_{t-1}, \dots ,x_{t-p}\}$ values and the moving average (MA) model that defines $x_{t}$ based on the linear combination of the past $q$ random perturbations to the process mean. The ``I/integrated'' part of the name refers to the process of $d$-order differencing, a method for approximating constant mean and, to a lesser extent, autocovariance, through time; the simplest and most common example of this is first-order differencing which is concerned with modeling the change in $x_{t}$ ($\Delta x_{t} = x_{t} - x_{t-1}$) instead of $x_{t}$ directly. Thus, an ARIMA model is defined by 3 parameters: $p$, $d$, and $q$.

This work examined two baseline models related to the commonly used ARIMA model\textemdash more precisely, with both being variants of the AR model. All neural network architectures examined in this work utilize $6$ previous scenes to forecast the next scene (unless stated otherwise), so the AR($p=6$) model is an intuitive baseline analog. This can be expressed mathematically as:

\small
\begin{equation}
    x_{t} = \sum_{i=1}^{6} \phi_{i}x_{t-i}
\end{equation}
\normalsize

Likewise, scientifically, the move naive forecast is to assume no change from the most recent scenes, so this will also be examined by deploying an AR($p=1$) model with $\phi_{1} = 1$.

Lastly, a final baseline model is inspired from the domain's data processing\textemdash predicting all $0$ values as the forecast. In light of the data processing, this forecast can be interpreted as predicting a scene of homogeneous temperature with no pixels exceeding the background average. 

\subsubsection{Existing neural network architectures}

Among neural network architectures, the long short-term memory (LSTM) model is very popular. Like its preceding recurrent neural network (RNN), this model is designed for sequential (rather than independent) data, and it expanded upon its predecessor by introducing gate units that are trained to learn the optimal flow of new data and retention of prior information. The equations of this model are given below:

\small
\begin{equation}
    i_{t} = \sigma(W_{xi}x_{t}+W_{hi}h_{t-1}+b_{i})
\end{equation}
\begin{equation}
    f_{t} = \sigma(W_{xf}x_{t}+W_{hf}h_{t-1}+b_{f})
\end{equation}
\begin{equation}
    c_{t} = f_{t}\odot c_{t-1}+i_{t}\odot tanh(W_{xc}x_{t}+W_{hc}h_{t-1}+b_{c})
\end{equation}
\begin{equation}
    o_{t} = \sigma(W_{xo}x_{t}+W_{ho}h_{t-1}+b_{o})
\end{equation}
\begin{equation}
    h_{t} = o_{t}\odot tanh(c_{t})
\end{equation}
\normalsize

\noindent Here, $\circ$ denotes the Hadamard product for element-wise multiplication of two matrices, $\sigma$ represents the sigmoid activation function, lower case variables indicate scalars or vectors, and upper case variables indicate matrices. Likewise, these notation rules are adapted throughout the section, modifying original equations as needed. 

The traditional LSTM was modified by Shi et al. \cite{Shi2015} to perform precipitation nowcasting on radar maps by changing both the inputs and the hidden states from scalars/vectors to matrices and applying the learned weights via the convolution ($\ast$) or ``sliding window'' operator in Equations 2-5:

\small
\begin{equation}
    I_{t} = \sigma(W_{xi} \ast X_{t}+W_{hi} \ast H_{t-1}+b_{I})
\end{equation}
\begin{equation}
    F_{t} = \sigma(W_{xf} \ast X_{t}+W_{hf} \ast H_{t-1}+b_{F})
\end{equation}
\begin{equation}
    C_{t} = F_{t}\odot C_{t-1}+I_{t}\odot tanh(W_{xc} \ast X_{t}+W_{hc} \ast H_{t-1}+b_{C})
\end{equation}
\begin{equation}
    O_{t} = \sigma(W_{xo} \ast X_{t}+W_{ho} \ast H_{t-1}+b_{O})
\end{equation}
\begin{equation}
    H_{t} = O_{t}\odot tanh(C_{t})
\end{equation}
\normalsize

\noindent The result is the now-popular convolutional LSTM with proven results on scientific raster time series, and the architecture has since been used to predict future temperature \cite{Xiao2019}, precipitation \cite{Kim2017}, and land cover \cite{Stoian2019}.

Separately, Zhu et al. \cite{Zhu2017a} incorporated two time gates to allow the ``notion of interval'' or time between observations to improve content recommendation services (e.g., music and scientific literature). These time gates serve as further learned modulators to the flow of short-term and long-term memory, yielding the Time-LSTM:

\small
\begin{equation}
    t1_{t} = \sigma(W_{x1}x_{t}+\sigma(W_{t1}\Delta t_{t}) + b_{1})
\end{equation}
$$ \mbox{such that } W_{t1} \leq 0 $$
\begin{equation}
    t2_{t} = \sigma(W_{x2}x_{t}+\sigma(W_{t2}\Delta t_{t}) + b_{2})
\end{equation}
\begin{equation}
    \widetilde{c_{t}} =(1-i_{t}\odot t1_{t})\odot c_{t-1}+i_{t}\odot t1_{t}\odot \sigma(W_{xc}x_{t}+W_{hc}h_{t-1}+b_{c})
\end{equation}
\begin{equation}
    c_{t} =(1-i_{t})\odot c_{t-1}+i_{t}\odot t2_{t}\odot \sigma(W_{xc}x_{t}+W_{hc}h_{t-1}+b_{c})
\end{equation}
\begin{equation}
    h_{t} = o_{t}\odot tanh(\widetilde{c_{t}})
\end{equation}
\normalsize

\noindent Similarly, Baytas et al. \cite{Baytas2017} developed the Time-Aware LSTM that analyzed irregularly timed patient records to predict patient groups with similar disease progression pathways. Instead of using further nonlinear and learned gates, however, the Time-Aware LSTM simply reallocates the memory contributions of long-elapsed inputs from short-term to long-term memory proportional to the time elapsed via a monotonically non-increasing function $g$:

\small
\begin{equation}
    c^{S}_{t-1} = tanh(W_{d}c_{t-1}+b_{d})
\end{equation}
\begin{equation}
    \hat{c}^{S}_{t-1} = c^{S}_{t-1} \odot g(\Delta_{t})
\end{equation}
\begin{equation}
    c^{T}_{t-1} = c_{t-1}-c^{S}_{t-1}
\end{equation}
\begin{equation} 
    c^{\ast}_{t-1} = c^{T}_{t-1} + \hat{c}^{S}_{t-1}
\end{equation}
\begin{equation}
    \widetilde{c} = tanh(W_{xc}x_{t} + W_{hc}h_{t-1} + b_{c})
\end{equation}
\begin{equation}
    c_{t} = f_{t}\odot c^{\ast}_{t-1} + i_{t}\odot \widetilde{c}
\end{equation}
\normalsize

\noindent Thus, the Time-LSTM and Time-Aware LSTM largely modify and expand Equation (4) of the original LSTM \footnote{It is worth noting that the Time-LSTM defines $\Delta_{t}$ as the time following an observation ($t_{i+1} - t_{i}$), while the Time-Aware LSTM defines $\Delta_{t}$ as the time preceding an observation ($t_{i} - t_{i-1}$).}.

\subsubsection{Proposed neural network architectures}

There does, however, remain the opportunity to modify these equations to include the composite spatial handling of the convolutional LSTM and the irregular temporal handling of the Time-LSTM or Time-Aware LSTM, and this is somewhat straightforward due to their shared origin and minimal modifications of the traditional LSTM. I express these modifications below for (1) Time-LSTM + ConvLSTM:

\small
\begin{equation}
    I_{t} = \sigma(W_{XI} * X_{t} + W_{HI} * H_{t-1} + b_{I})
\end{equation}
\begin{equation}
    T1_{t} = \sigma(W_{XT1} * X_{t} + \sigma(W_{TT1} * \Delta T_{t}) + b_{T1})
\end{equation} 
$$ \mbox{such that } W_{TT1} \leq 0 $$ 
\begin{equation}
    T2_{t} = \sigma(W_{XT2} * X_{t} + \sigma(W_{TT2} * \Delta T_{t}) + b_{T2}) 
\end{equation}
\begin{equation}
\begin{split}
    \widetilde{C_{t}} = (1 - I_{t} \circ T1_{t}) \circ C_{t-1} + I_{t} \circ T1_{t} \circ \\ \sigma(W_{XC} * X_{t} + W_{HC} * H_{t-1} + b_{C})
\end{split}
\end{equation}
\begin{equation}
\begin{split}
    C_{t} = (1-I_{t}) \circ C_{t-1} + I_{t} \circ T2_{t} \circ \\ \sigma(W_{XC} * X_{t} + W_{HC} * H_{t-1} + b_{C})
\end{split}
\end{equation}
\begin{equation}
    O_{t} = \sigma(W_{XO} * X_{t} + W_{TO} * \Delta T_{t} + W_{HO} * H_{t-1}+b_{O})
\end{equation}
\begin{equation}
    H_{t} = O_{t} \circ \sigma(\widetilde{C_{t}})
\end{equation}
\normalsize

\noindent and (2) Time-Aware LSTM $+$ ConvLSTM: 

\begin{equation}
    I_{t} = \sigma(W_{XI} * X_{t} + W_{HI} * H_{t-1}+b_{I})
\end{equation}
\begin{equation} 
    F_{t} = \sigma(W_{XF} * X_{t} + W_{HF} * H_{t-1}+b_{F})
\end{equation}
\begin{equation}
    O_{t} = \sigma(W_{XO} * X_{t} + W_{HO} * H_{t-1}+b_{O})
\end{equation}
\begin{equation}
    C^{S}_{t-1} = tanh(W_{D} * C_{t-1} + b_{D})
\end{equation}
\begin{equation} 
    \hat{C}^{S}_{t-1} = C^{S}_{t-1} \circ g(\Delta T_{t})
\end{equation}
\begin{equation}
    C^{T}_{t-1} = C_{t-1} - C^{S}_{t-1}
\end{equation}
\begin{equation} 
    C^{*}_{t-1} = C^{T}_{t-1} + \hat{C}^{S}_{t-1}
\end{equation}
\begin{equation}
    \widetilde{C} = tanh(W_{XC} * X_{t} + W_{HC} * H_{t-1} + b_{C})
\end{equation}
\begin{equation}
    C_{t} = F_{t} \circ C^{*}_{t-1} + I_{t} \circ \widetilde{C}
\end{equation}
\begin{equation}
    H_{t} = O_{t} \circ tanh(C_{t})
\end{equation}
\normalsize

The above equations are minimally modified from their original forms, but, notably, the use of peephole connections in the original Time-LSTM were not included because (i) the exclusion yields a simpler model, and (ii) a seminal paper \cite{Greff2017} found that removing peephole connections from LSTMs did not significantly reduce performance. In both cases, the convolutional LSTM's C gate has been modified to incorporate the potentially nonlinear effect of the time interval ($\Delta T_{t}$) between the current observation ($X_{t}$) and the previous state ($H_{t-1}$) on the current state ($H_{t}$). Here, the time interval ($\Delta T_{t}$) is also a spatial structure that allows temporal specification per pixel; this enables the use of partially obstructed images which may be caused by different orbital paths and/or incomplete cloud obstruction. The current state ($H_{t}$) is naively a spatial structure, but it may be further processed to some metric of interest (e.g. maximum pixel value, the number of anomalous pixels) or used as inputs to further neural network layers.

\subsubsection{U-Net}

Manual inspection of early results for the convolution-based models suggested that these models tended to produce overly smoothed predictions. This can be seen as a problem with precise localization, which the U-Net architecture \cite{Ronneberger2015} provides a solution for. The U-Net architecture was designed for image segmentation with no intention for sequential tasks. In the original U-Net implementation, a deep convolutional neural network is improved by creating symmetrical hidden dimensions which start relatively low then increase and return to the original specifications (e.g., $64$, $128$, $64$); as this happens, the spatial size shrinks then expands while propagating the outputs of earlier layers into the later but symmetrical layers (i.e., $64_{1}$'s output propagates to $64_{2}$'s input) to promote localization and improved resolution. Here, we implement the symmetrically sized hidden dimensions and the propagation of earlier symmetrical layers' output to the convolution-based model with the lowest validation set RMSE.

\subsection{Optimization Details}

The data set was separated with a $70/15/15$ split chronologically by each volcano into a training, validation, and test set. The training set was used to fit model parameters, the validation set was used to compare different models as well as compare the same model fit with different regularization strengths, and the test set was aside for evaluation of the final, best model. Every model was fit with $5$ different regularization strengths ranging from $0.0001$ to $1$ by a factor of $10$; higher values of this parameter help reduce overfitting to the training set by penalizing large model parameters.

Choosing the exact number of hidden layers as well as the number of units/neurons in each layer is not straightforward, with the only general rules being that more parameters increase performance, computational burden, and the need to combat overfitting. The exact hidden layer choices used in this project were taken from the published best results of the existing neural networks used here \cite{Shi2015, Zhu2017a, Baytas2017}. All models were trained for 100 epochs with the Adam optimizer \cite{Kingma2015} to minimize mean squared error. Optimization occurred in varying mini-batch sizes that were determined to maximize GPU utilization.

All models were implemented in the Python programming language via the PyTorch machine learning framework. The hardware used for this project was an HPC cluster with 10 NVIDIA Tesla V100 GPUs with 32 gigabytes of RAM each; all neural network computations used this hardware in parallel.

\section{Results and Discussion}

\subsection{Experiment 1: Best Model for Forecasting Images} 

\textbf{Table 1} presents the validation set RMSE along with important training details for all the models. Among the naive methods, using the most recent scene as the forecast for the next scene led to an RMSE of $5.291^{\circ}$C while, alternatively, assuming a homogenous temperature landscape (``All Zeros'') led to an RMSE of $4.738^{\circ}$C. Representing the only learned statistical model, the AR(6) returned an RMSE of $4.322^{\circ}$C. Among the non-convolutional neural networks, the standard LSTM models returned RMSEs ranging from $4.217$ to $4.265^{\circ}$C, the Time-LSTM models returned RMSEs ranging from $4.694$ to $4.870^{\circ}$C, and the Time-Aware LSTM models returned RMSEs ranging from $4.660$ to $4.726 ^{\circ}$C. The latter two converged very closely to the All Zeros model, returning low magnitude predictions in the best results. Convolution-based neural networks returned RMSEs between $4.164$ and $4.428^{\circ}$C with less variance than the non-convolutional neural networks. The lowest RMSE was achieved by the proposed ConvLSTM + Time-LSTM + U-Net architecture. It was approximately $10$ times faster than the lowest RMSE LSTM and required approximately $40$ times fewer parameters.

\begin{table*}[t]
\caption{Validation set performance and training details for all models}
\centering
\begin{tabular}{c|l|l|l|c|c}
Model Type & Model & Hidden Dim.s & RMSE ($^{\circ}$C)& Number of Parameters & Training Time (hours) \\
\hline
Naive & Last Scene & -- & 5.291 & -- & -- \\ 
& All Zeros & -- & 4.738 & -- & -- \\
\hline
Statistical & AR(6) & -- & 4.322 & 6 & 0.10 \\
\hline
Non-spatial NN & LSTM & 128 & 4.265 & 67080 & 0.09 \\
& & 1028, 512 & 4.229 & 7393384 & 2.05 \\
& & \textbf{2000, 2000} & \textbf{4.217} & \textbf{48032008} & \textbf{4.91} \\
& Time-LSTM & 128 & 4.870 & 51213 & 0.29 \\
& & 1028, 512 & 4.706 & 5555825 & 4.00 \\
& & 2000, 2000 & 4.694 & 36052013 & 10.10 \\
& Time-Aware LSTM & 128 & 4.660 & 83594 & 0.25 \\
& & 1028, 512 & 4.719 & 8713854 & 4.15 \\
& & 2000, 2000 & 4.726 & 56036010 & 11.37 \\
\hline
Spatial NN & ConvLSTM & 64, 64 & 4.230 & 447528 & 0.26 \\
&  & 128, 64, 64 & 4.428 & 1335080 & 0.41 \\
& ConvLSTM + Time-LSTM & 64, 64 & 4.255 & 341967 & 0.31 \\
&  & 128, 64, 64 & 4.250 & 1013903 & 0.46 \\
& ConvLSTM + Time-Aware LSTM & 64, 64 & 4.256 & 521394 & 0.31 \\
&  & 128, 64, 64 & 4.394 & 1556530 & 0.45 \\
& ConvLSTM + Time-LSTM + U-Net & \textbf{64, 128, 64} & \textbf{4.164} & \textbf{1233359} & \textbf{0.53} \\
\end{tabular}
\end{table*}

For the best model of each type (i.e., statistical, non-convolutional, and convolutional), the models were separately trained for $1000$ epochs to determine if further training resulted in better performance, but, instead, it was found that validation set RMSE increased for all models with this extended training. For the AR(6), validation set RMSE increased from $4.322$ to $4.357^{\circ}$C; for the LSTM, $4.217$ to $4.293^{\circ}$C; and, for the ConvLSTM + Time-LSTM + U-Net, $4.164$ to $4.182^{\circ}$C. The models originally trained for $100$ epochs are thus used for subsequent discussion, and the three mentioned in this paragraph will be referred to as the ``selected models''.

Predictions for the lowest RMSE model are presented in \textbf{Figure 6} and \textbf{Figure 7}. Both of these demonstrations compare observations with predictions during large effusive events. In \textbf{Figure 6} the predictions are well aligned with the observations. It can be noted that the predicted sequence failed to predict new spatial features until after they emerged in observations. Interestingly, as the lava flow continues, the model emphasized particular hotspot areas that were present in prior scenes while broadening the width of more linear features. While \textbf{Figure 7} is also a large lava flow, its data are much less ideal\textemdash note the persistent presence of partial cloud cover as well as missing edge values which required filling with previous scenes. Despite these obstructing conditions, the model continued to predict the presence of a large lava flow in the correct general direction and shape.

\begin{figure*}[t]
\centering
\includegraphics[width=\linewidth]{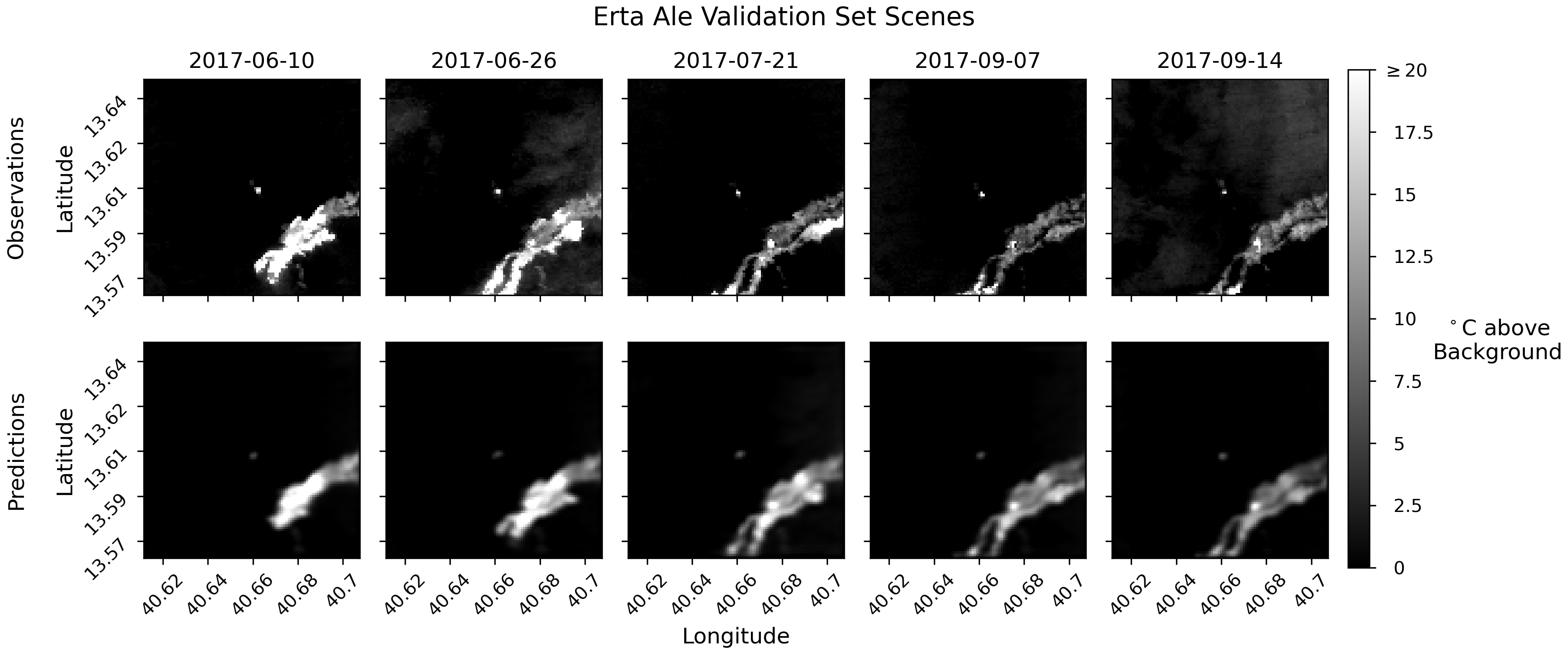}
\caption{Demonstration of validation set predictions for the best model. Top row: observations. Bottom row: predictions.}
\end{figure*}

\begin{figure*}[t]
\centering
\includegraphics[width=\linewidth]{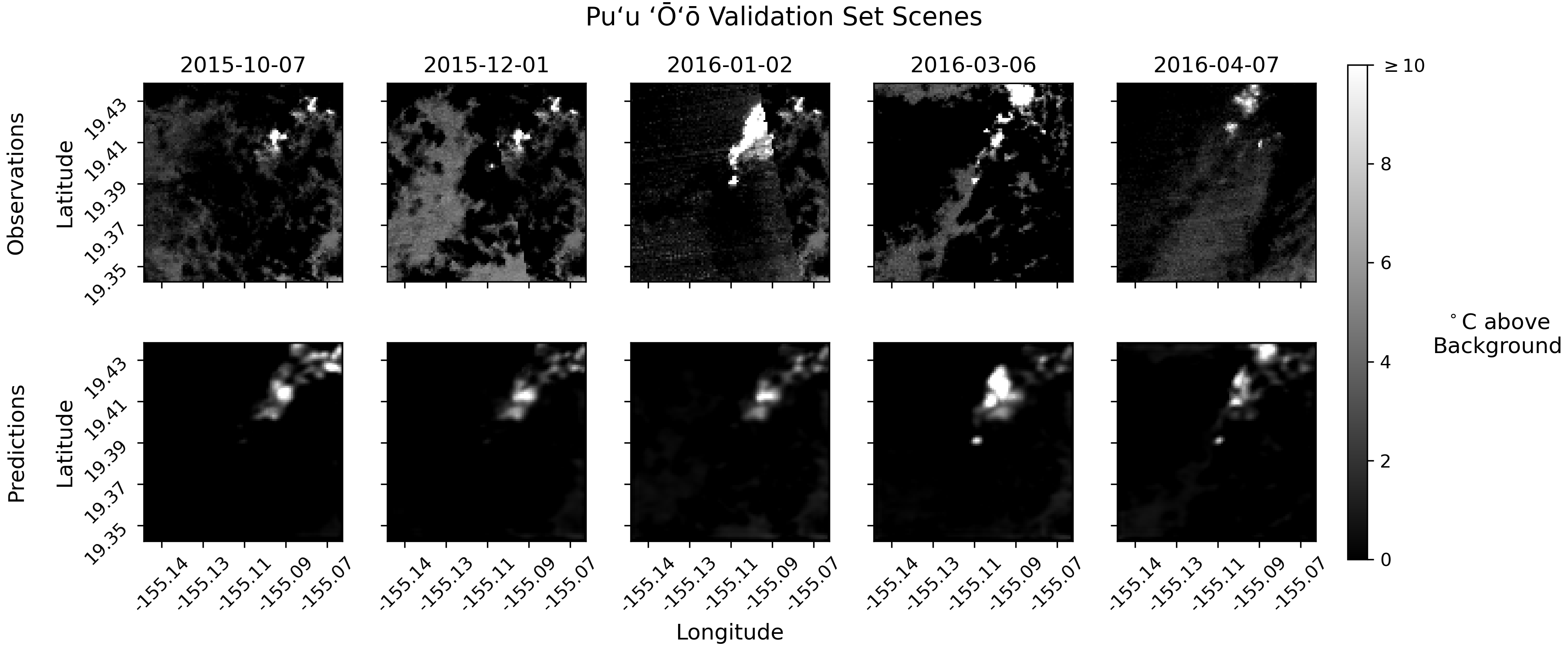}
\caption{Demonstration of validation set predictions for the best model. Top row: observations. Bottom row: predictions. Note the cloudy conditions and missing values that have been filled (as apparent by the diagonal boundaries in the observations for 2015-12-01 and 2016-01-02).}
\end{figure*}

The distributions of predicted pixel values for the selected models as well as a poor-performing model (Time-LSTM) were examined and are presented in \textbf{Figure 8}. Barring the poor-performing model, it was found that the models with lower RMSE tended to predict lower-valued extremes than higher RMSE models and all models predicted lower-valued extremes than the true observations.

\begin{figure}[t]
\centering
\includegraphics[width=\linewidth]{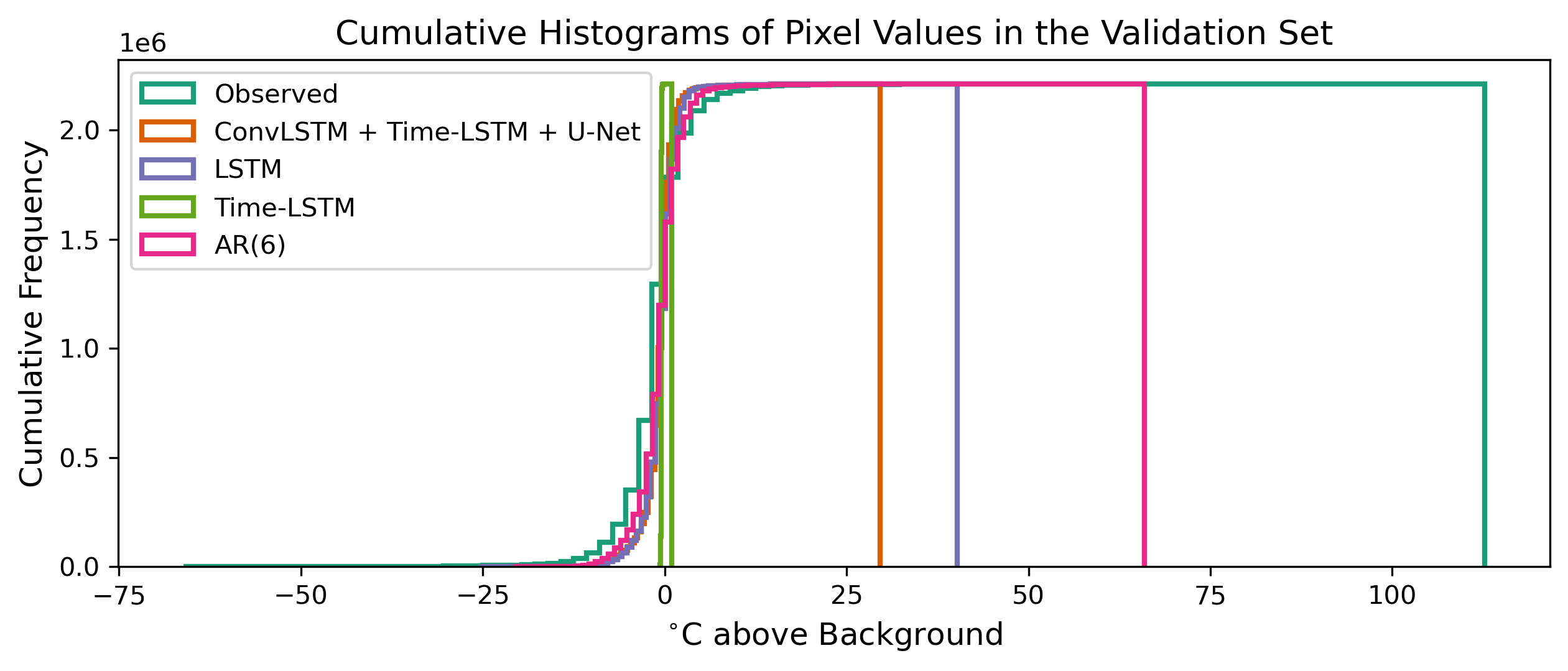}
\caption{Cumulative histograms of pixel values in the validation set for the observations and selected models. Vertical lines indicate maximums.}
\end{figure}

All in all, the explicit modeling/optimization task was to predict image sequences that are irregularly spaced in time with minimal RMSE relative to the true observations. Each model had different characteristics which influenced its ability to accomplish this task. For example, the AR(6) had very few parameters while some of the neural networks had millions of parameters; the Time-LSTM considered the time between scenes while the ConvLSTM considered a pixel's spatial neighbors. Ultimately, the proposed model, specifically the ConvLSTM + Time-LSTM + U-net, considered more characteristics of the data (e.g., spatial, irregular in time, and high resolution) and accomplished this explicit goal the best. It is worthwhile to note that while all models have low error relative to the full range of the data that these error differences are spread across $2211840$ pixels\textemdash $240$ images in the withheld validation set, each of which contains $9216$ pixels.

\subsection{Experiment 2: Derived Time Series}

With infrared-based imagery, there are many derived time series of interest in volcanic monitoring. Specifically, with temperature data, radiance-based measurements cannot be examined, so here we focus on (1) maximum temperature above the background, (2) the number of hotspots, and (3) distance of furthest hotspot from the summit. All derived time series are concerned with anomalous or extreme temperatures, so it was expected that the models' inability to predict the true scale of extreme temperatures would inhibit derived time series. To alleviate this, predicted time series were derived both before and after histogram matching the validation set predictions to the training set observations. A demonstration of this can be seen in \textbf{Figure 9}.

\begin{figure}[t]
\centering
\includegraphics[width=\linewidth]{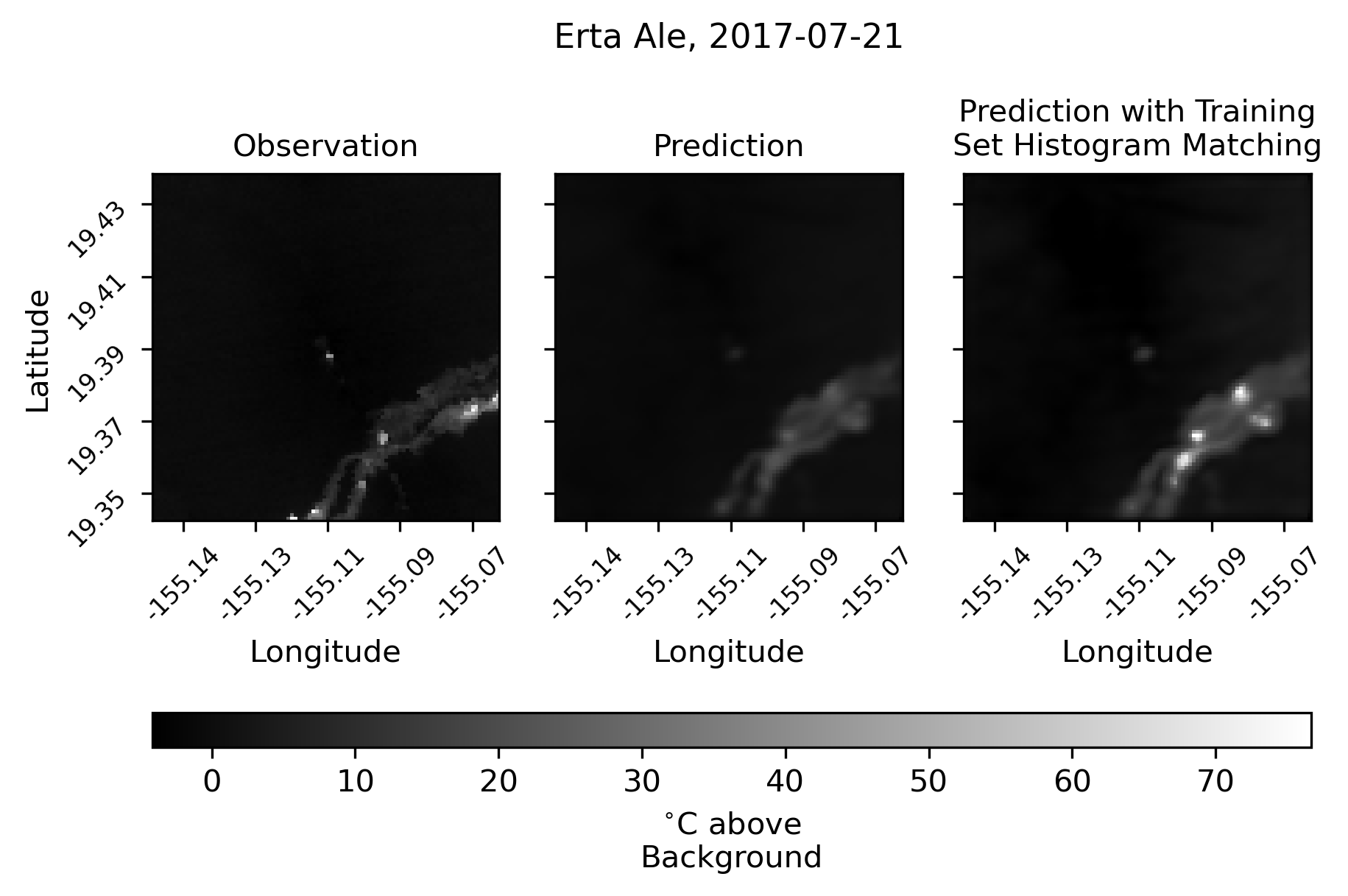}
\caption{A demonstration of histogram matching a validation set prediction to the training set. Left: observation. Middle: prediction. Right: prediction with histogram matching. The same color scale is used for all three images. Note the increased abundance of observation-scale extremes after histogram matching to the training set.}
\end{figure}

\textbf{Table 2} contains the error of the selected models as well as the poor-performing Time-LSTM on the derived time series. See \textbf{Figure 10} for a demonstration of these time series using the validation set, histogram matching, and the ConvLSTM + Time-LSTM + U-Net model.

\begin{table*}[t]
\caption{Derived Time Series Metrics for Selected Models}
\centering
\begin{tabular}{c|c|c|c|c}
Model & Histogram & Maximum & Number of & Maximum Distance of a Hotspot (m) \\
& Matching & Temperature ($^{\circ}$C) & Hotspots & from the Summit \\
\hline
AR(6) & No & 33.304 & 20.553 & 1190.338 \\
AR(6) & Yes & 32.255 & 18.180 & 1067.456 \\
\hline
LSTM & No & 40.604 & 25.249 & 1421.756 \\
LSTM & Yes & \textbf{32.120} & 18.303 & 1105.383 \\
\hline
Time-LSTM & No & 57.930 & 25.347 & 1513.082 \\
Time-LSTM & Yes & 32.361 & \textbf{17.468} & \textbf{825.800} \\
\hline
ConvLSTM + Time-LSTM + U-net & No & 46.262 & 25.347 & 1513.082 \\
ConvLSTM + Time-LSTM + U-net & Yes & 36.642 & 17.884 & 1090.002 \\
\end{tabular}
\end{table*}

\begin{figure*}[t]
\centering
\includegraphics[width=\linewidth]{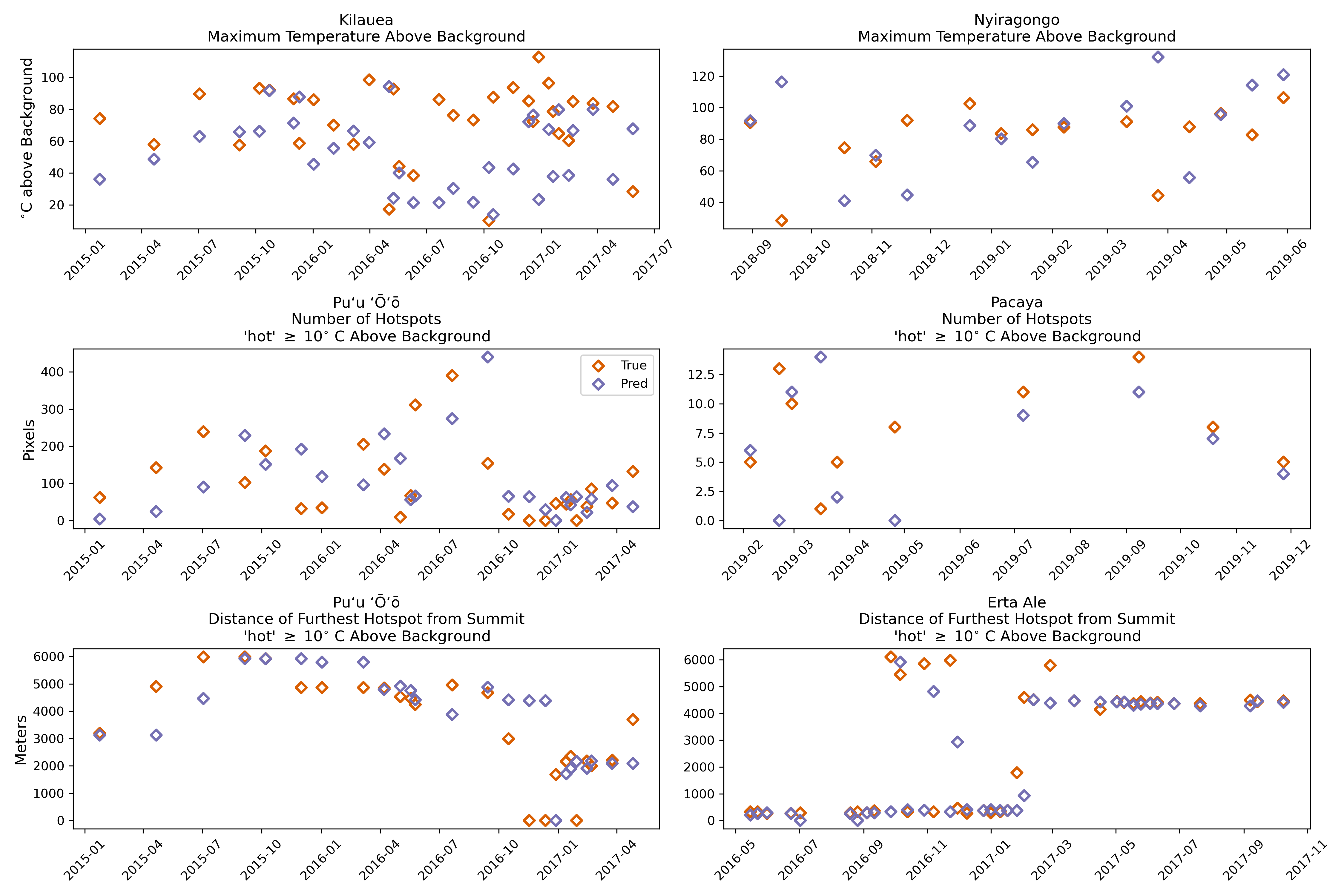}
\caption{A demonstration of derived time series using validation set data from the best model. Included here are time series for maximum temperature above the background, the number of hotspots, and distance of furthest hotspot from the summit. ``Hot'' is defined as $10^{\circ}$C above the background, and volcanoes plotted here include Kīlauea, Nyiragongo, Pu'u 'Ō'ō, Pacaya, and Erta Ale.}
\end{figure*}

Before histogram matching, the AR(6) achieves the lowest RMSE on all three derived time series. This is attributable to the fact that it predicts the largest maximum values of the available models. After histogram matching, which accounts for the apparent tendency of neural networks to produce low-scale extremes, the AR(6) model is no longer the best at any derived time series. Likewise, the proposed model (ConvLSTM + Time-LSTM + U-net), which achieved the lowest RMSE on predicting the image sequences was unable to reproduce any derived time series with the lowest RMSE. Instead, after histogram matching, the traditional LSTM achieved the lowest RMSE recreating the maximum temperature above the background ($32.120^{\circ}$C), while the Time-LSTM achieved the lowest RMSE on the number of hotspots ($17.468$ pixels) and the maximum distance of a hotspot from the summit ($825.800$ meters).

This discrepancy is interesting because the Time-LSTM was included in further comparisons since its best optimization resulted in the worst validation set image sequences (RMSE = $4.694^{\circ}$C). Further examination revealed that this is because it tended to converge to near-zero image predictions (minimum = $-0.907^{\circ}$C, maximum = $0.958^{\circ}$C). Together, this suggests that despite struggling immensely with scale, the Time-LSTM learned superior pattern recognition within a very narrow comfort zone for this task that was indirectly optimized for.

Scientifically, this suggests an important implication\textemdash when forecasting multipurpose data with many potential uses, directly minimizing error on the data does not necessarily minimize error on its potential uses or derivations. Moving forward, it would be best to explicitly account for these other metrics directly in the loss function (i.e., optimization problem). Compared to highly engineered, domain-based algorithms, deep learning models are relatively simple and rely on heavy parameterization along with large amounts of data to improve performance. With immense interest in the value of deep learning for scientific tasks, it will likely be a very promising pursuit to engineer loss functions with the multiple scientific priorities and constraints in mind\textemdash an example of this is can be seen when forecasting water temperature constrained by physical laws \cite{Read2019}.

\subsection{Experiment 3: Training with Individual Volcanoes}

Broadly, RMSE decreased when the training set was limited to individual volcanoes (\textbf{Table 3)}. In general, volcanoes with fewer data produced worse results, but performance did not increase exclusively with training set quantity. For example, training exclusively with Erta Ale or Kīlauea data led to lower RMSE than training exclusively with Erebus data ($4.356^{\circ}$C, n = 178 and $4.390^{\circ}$C, n = 150 versus $4.684^{\circ}$C, n = 336), and training exclusively with Nyiragongo data led to higher RMSE than training exclusively with Masaya data ($5.365^{\circ}$C, n = 74 versus $5.343^{\circ}$C, n = 44). 

\begin{table*}[t]
\caption{Validation set performance for different training sets: all volcanoes or specific volcanoes}
\centering
\scalebox{0.92}{
    \begin{tabular}{c|c|c|c|c|c|c|c|c|c|c|c}
     & & \multicolumn{9}{c}{Validation Set Performance ($^{\circ}$C)} \\
    Training Volcano & Number of Sequences & All & Erebus & Erta Ale & Etna & Kīlauea & Masaya & Nyamulagira & Nyiragongo & Pacaya & Pu'u 'Ō'ō \\
    \hline
    All         & 1127 & 4.164 & 3.311 & 5.875 & 2.631 & 2.583 & 3.458 & 5.019 & 7.423 & 2.611 & 3.890 \\
    Erebus      & 336 & 4.684 & 3.959 & 6.160 & 4.629 & 2.880 & 5.532 & 5.369 & \textbf{7.100} & 3.052 & \textbf{3.884} \\
    Erta Ale    & 178 & 4.356 & 3.897 & \textbf{5.534} & 4.451 & 2.742 & \textbf{3.141} & \textbf{4.868} & \textbf{6.691} & 3.085 & 3.890 \\
    Etna        & 132 & 5.696 & 5.629 & 7.341 & 3.667 & 4.533 & 6.369 & 5.898 & 7.652 & 3.347 & 5.206 \\
    Kīlauea     & 150 & 4.390 & 3.863 & \textbf{5.486} & 4.737 & 2.729 & \textbf{3.229} & \textbf{4.955} & \textbf{6.745} & 3.350 & \textbf{3.882} \\
    Masaya      & 44 & 5.343 & 4.772 & 6.671 & 5.362 & 3.739 & 6.073 & 5.966 & 7.527 & 3.992 & 4.585 \\
    Nyamulagira & 53 & 5.345 & 4.846 & 6.727 & 5.122 & 3.785 & 6.074 & 5.920 & 7.508 & 3.838 & 4.620 \\
    Nyiragongo  & 74 & 5.365 & 4.788 & 6.678 & 5.422 & 3.765 & 6.103 & 6.001 & 7.552 & 3.970 & 4.611 \\
    Pacaya      & 50 & 5.453 & 5.103 & 6.919 & 4.717 & 4.011 & 6.166 & 5.915 & 7.540 & 3.736 & 4.802 \\
    Pu'u 'Ō'ō       & 120 & 4.946 & 4.340 & 6.356 & 4.967 & 3.119 & 5.636 & 5.469 & \textbf{7.241} & 3.543 & 4.109 \\
    \end{tabular}
}
\end{table*}

It is commonly noted that every volcano is different and, as such, patterns learned from one volcano may not generalize to another. It could have been possible that a volcano's validation set performance was minimized by training only with data from that volcano, but that was not found to be the case; in general, training with a specific volcano did not reduce RMSE on that volcano itself. The only exception to this was Erta Ale (all volcanoes RMSE = $5.875^{\circ}$C, Erta Ale only RMSE = $5.534^{\circ}$C). The reasons for this are not known; it may be that patterns from the other volcanoes do not apply well to Erta Ale, and/or the fact that, unlike most volcanoes in this data set, Erta Ale has relatively flat, cloud-free terrain.

In $8/9$ cases, a training set consisting of one volcano did not benefit validation set performance on that volcano itself, but in $4/9$ cases, this did reduce validation set RMSE on at least one other volcano. This was the case when the training set consisted of only Erebus, Erta Ale, Kīlauea, and Pu'u 'Ō'ō resulting in validation set RMSE reductions in $2/8$, $3/8$, $5/8$, and $1/8$ other volcanoes, respectively. Among these, the model fit using only Kīlauea data improved validation set performance on at least the same volcanoes that the others did (i.e., training sets of Erebus, Erta Ale, and Pu'u 'Ō'ō). Kīlauea is one of the most well-studied volcanoes in the world and, fortunately, this appears to suggest that thermal patterns learned from it may generalize better than other well-known basaltic volcanoes.

\subsection{Opportunities}

A forefront weakness in this forecasting project is that high-resolution thermal imagery is the only data considered. Even within thermal data sources, there are additional data sources that could be integrated\textemdash most obviously MODIS. Additionally, regional weather satellites such as GOES could prove valuable and would still provide nighttime imagery. Beyond nighttime imagery, there are also the options of Landsat and Sentinel-2. With additional expertise it would also be beneficial to include other forms of remote sensing, such as InSAR for ground deformation and IR/UV for gas emissions, and, when available, performance would likely benefit from the inclusion of ground monitoring efforts. Ultimately, volcanoes display a variety of behavior that can be measured on the ground and from space; here, we focused on thermal satellite imagery, one of the most globally available and widely adopted.

Moving forward, the task of integrating all of this data will likely not be straightforward. Experts in the field have speculated that ''few developments promise to revolutionize eruption forecasting — along with so many other branches of science — as does machine learning.'' \cite[p. 23]{Poland2020}. We argue here that important developments moving forward will likely require the development of machine/deep learning methods with scientific goals and data characteristics in mind. Geoscience fields, for example, are generally data-sparse but rich with theory-based models \cite{Karpatne2019}, and it has been found that integrating scientific theory with machine learning helps alleviate the burden of low data quantity \cite{Read2019}. Additionally, scientific data may not have the same characteristics as the data that machine learning methods were developed for. For example, Earth data are often irregular in time and space, and imagery may be very different (e.g., gravity-based data, interferograms, hyperspectral) from the typical optical situations that computer vision is concerned with. In summary, there are many broad directions and considerations for machine learning in science, and scientific discovery and application will likely benefit from models that are creatively specified to not only the data but also the governing scientific goals and principles.

\section{Conclusion}

Here we present one of the first projects forecasting volcanic activity with machine learning. Specifically, we forecasted ASTER nighttime temperature imagery for 9 volcanoes between $1999$ and $2020$. To better match the remote sensing data characteristics (i.e., temporally irregular image sequences), we proposed three new neural network architectures, and among them, one achieved lower RMSE than all considered models (e.g., autoregression, LSTM, ConvLSTM). All models struggled with scale, but it was found that histogram matching alleviated this, producing derived time series (e.g., maximum temperature above background, the number of hotspots, and distance of furthest hotspot from the summit) that generally approximated reality. It was discovered, however, that the lowest RMSE on forecasting future scenes did not ensure the lowest RMSE on derived tasks. This suggests a need for science-driven optimization targets, especially when forecasting multipurpose monitoring streams. Lastly, it appears that patterns learned from multiple volcanoes generally increase forecasting performance relative to single-volcano training sets. The pattern of improved validation set performance was not exclusively tied to data set size, suggesting that certain volcanoes provide generalizable patterns despite small data sets. Interestingly, the models fit to Kīlauea, one of the most well-studied volcanoes in the world, were found to generalize better than any other volcano considered here\textemdash further reducing RMSE relative to the entire data set in $5/9$ volcanoes.

This work demonstrates the value of implementing data-driven changes to popular neural networks for remote sensing forecasting, and we expect that it should generalize well to other Earth observation tasks with data that are irregularly available through time. This also demonstrated a weakness in using standard optimization targets (e.g., base data RMSE) for multipurpose scientific data; performing best at predicting data does not necessarily indicate the best performance on the final uses of that data. Machine learning has the potential for transformative results on scientific tasks, but it is paramount that these efforts make data-driven adjustments, consider the full problem to be explicitly optimized, and integrate existing scientific theory/principles.

\appendices

\section{Test Set Performance}

Many important findings in this paper revolved around training conditions (specifically, comparing different types of models and models that utilized different subsets of the available data). As such, the test set\textemdash sat aside for final performance evaluation of the best model\textemdash could not play a role in these narratives. Here, we present some test set findings when using the proposed model (ConvLSTM + Time-LSTM + U-net) that achieved the lowest validation set RMSE on the image forecasting task.

First, this model was fit on a combination of the training and validation sets (i.e., the chronologically first 85\% of scenes per volcano) to more fully utilize the available data. The resulting model was then evaluated on the test set. Overall, this model achieved a test set RMSE of $4.100^{\circ}$C\textemdash lower than all previously found validation set RMSEs (see \textbf{Table 1}).

In line with the discussions found in \textbf{Section 3, Subsections A} and \textbf{B}, this model refused to produce very anomalous temperatures\textemdash the maximum temperature it predicted for the entire test set was $10.268^{\circ}$C above the background, while the maximum observation for these scenes was $104.789^{\circ}$C above the background. This was problematic for the derived series where it achieved RMSEs of $46.718^{\circ}$C, $635.387$ pixels, and $3420.979$ meters for forecasting the maximum temperature above the background, the number of hotspots, and the maximum distance of a hotspot from the summit, respectively. Histogram matching to the training and validation set alleviated performance at forecasting the maximum temperature above the background, reducing the RMSE to $30.202^{\circ}$C; however, performance remained poor at forecasting the number of hotspots ($644.899$ pixels) and the maximum distance of a hotspot from the summit ($3819.775$ meters). In these tasks, stretching the small range of predicted values resulted in far too many hot pixels whereas before there were too few. For reference these values can be compared to \textbf{Table 2}, but it should be noted that those metrics were computed for different temporal ranges of the data (i.e., mutually exclusive dates between $2015$ and $2019$ versus dates between $2017$ and $2020$, depending on the volcano).

One of the events of particular interest in the test set is the 2018 eruptive activity experienced at Kīlauea and its East Rift Zone \cite{Neal2019}\textemdash represented in part by the Pu'u 'Ō'ō scenes in our data set. Unfortunately, no available scenes collected between May and August were deemed viable or uncertain, but in relation to our data set, this data would have been at the Pu'u 'Ō'ō and Kīlauea subsets, not the highly active lower East Rift Zone. There is, however, relatively dense data availability prior to May 2018.

Leading up the May 2018 lower East Rift Zone eruptions, the model does produce two interesting patterns in its forecasts (that have been histogram matched to the training and validation set to account for known issues in maximum values). The first is a large increase then sharp decrease in maximum temperature above the background at the Kīlauea summit (\textbf{Figures 11} and \textbf{12}); while this did not greatly mimic observations at the time, it did shortly proceed the May 2018 lava lake withdrawal that had consequences in local seismic activity and was indicative of broader changes in the plumbing system. It may be argued that this reduced temperature is an artifact of the preceding cloud cover, but \textbf{Figure 7} demonstrates that cloud cover alone is not sufficient to prohibit active forecasts. The second interesting forecast pattern is that the model predicted an increasing amount of hot spots at the Pu'u 'Ō'ō subset (\textbf{Figures 11} and \textbf{13}) leading up the lower East Rift Zone activity that stood in contrast to the fluctuating\textemdash if not declining\textemdash pattern seen in observations. In reality, the Pu'u 'Ō'ō vent ultimately collapsed, thus displaying no effusive hotspots; however, the ultimate result increased eruptive activity in the lower East Rift Zone. In summary, derived time series from the model forecasted sharply decreasing anomalous temperatures at Kīlauea's summit shortly before the lava lake withdrawal and increasingly wide eastward lava flows from Pu'u 'Ō'ō prior to Pu'u 'Ō'ō's vent collapse and the large lava flows in the lower East Rift Zone. 

\begin{figure*}[t]
\centering
\includegraphics[width=0.45\linewidth]{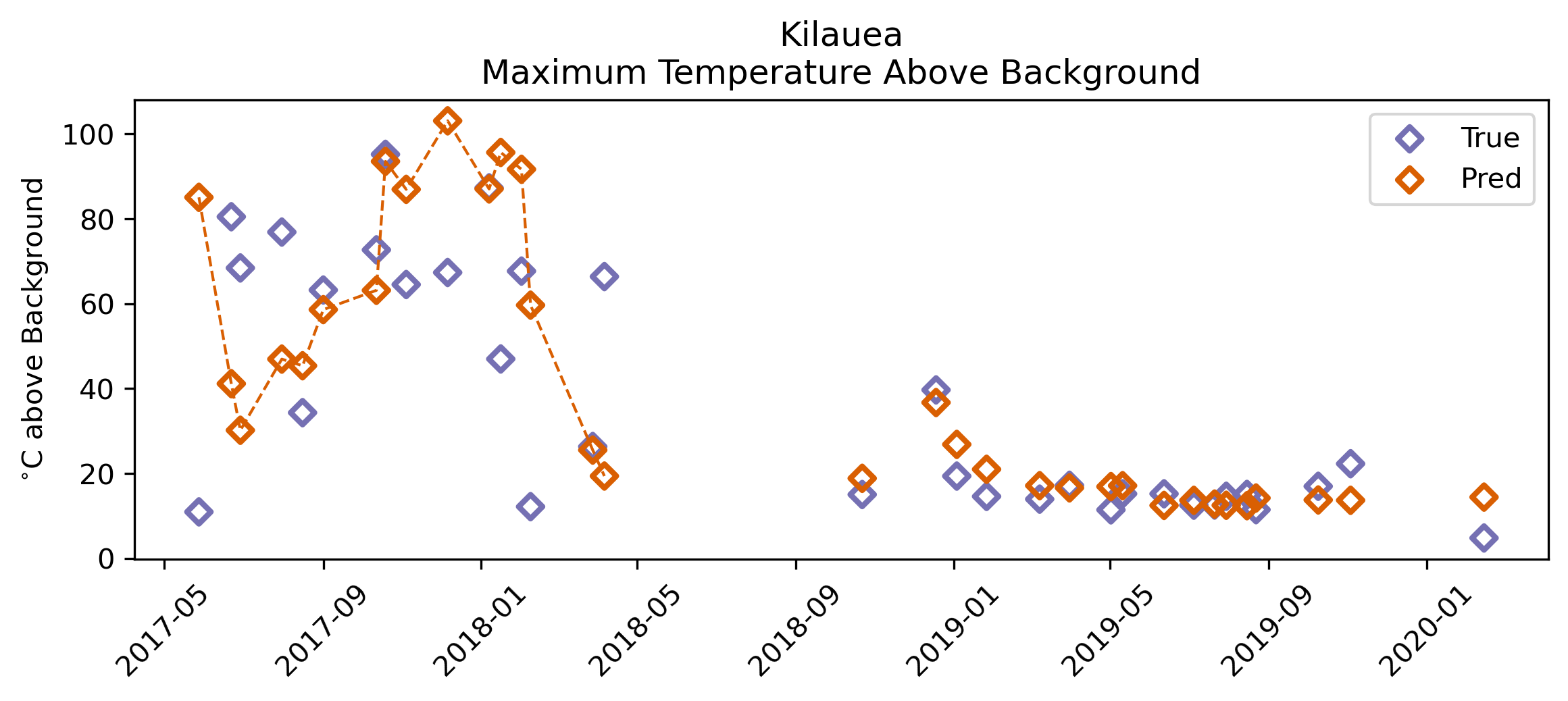}
\includegraphics[width=0.45\linewidth]{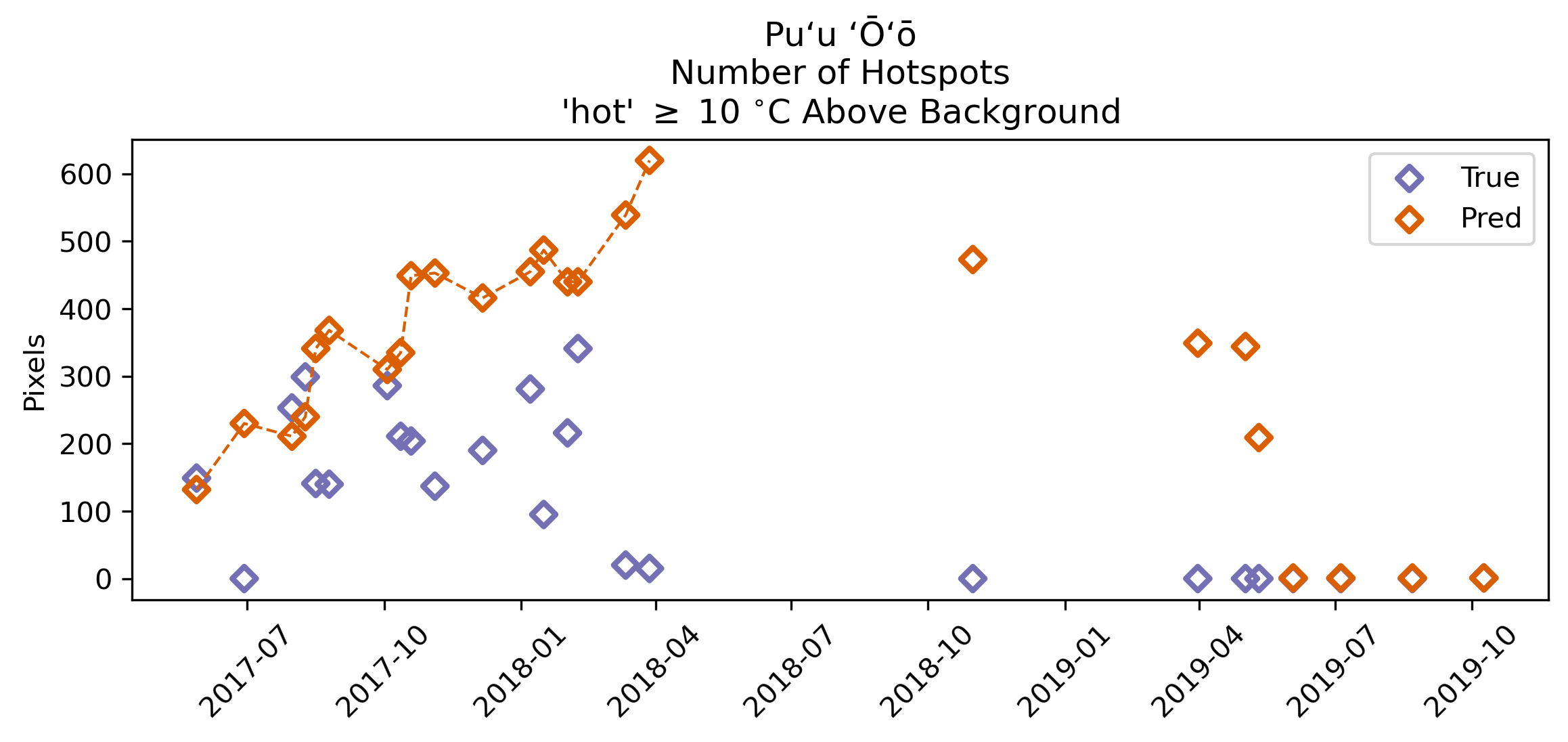}
\caption{Test set time series for maximum temperature above the background for Kīlauea's summit and number of hotspots for Pu'u 'Ō'ō. The dashed lines indicate the segment of forecasted values prior to the May 2018 lower East Rift Zone lava flows. Notably, the Pu'u 'Ō'ō vent collapsed on April 30, 2018, the Kīlauea lava lake began withdrawing on May 1, 2018, and there were large lava flows from multiple fissures in the lower East Rift Zone in May 2018.}
\end{figure*}

\begin{figure*}[t]
\centering
\includegraphics[width=\linewidth]{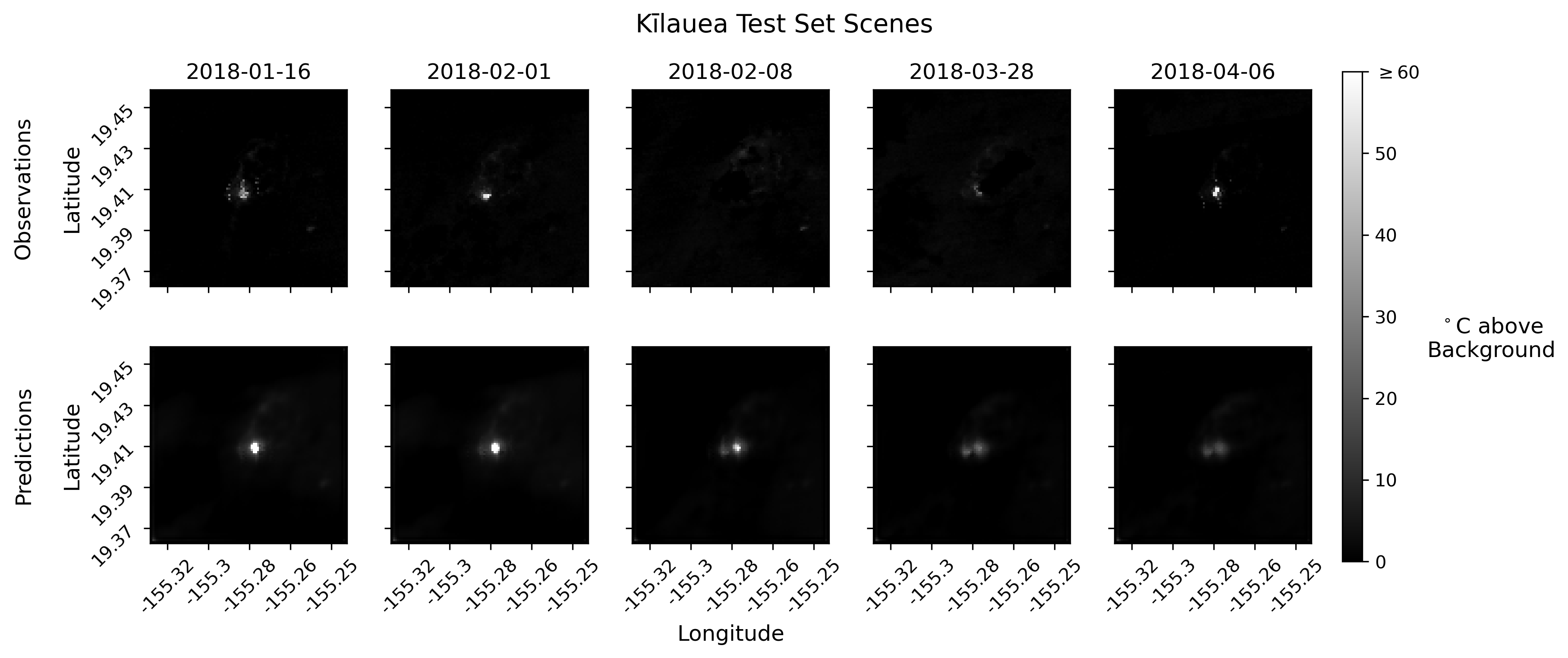}
\caption{Test set image forecasts for Kīlauea's summit. Notably, the lava lake began withdrawing on May 1, 2018.}
\end{figure*}

\begin{figure*}[t]
\centering
\includegraphics[width=\linewidth]{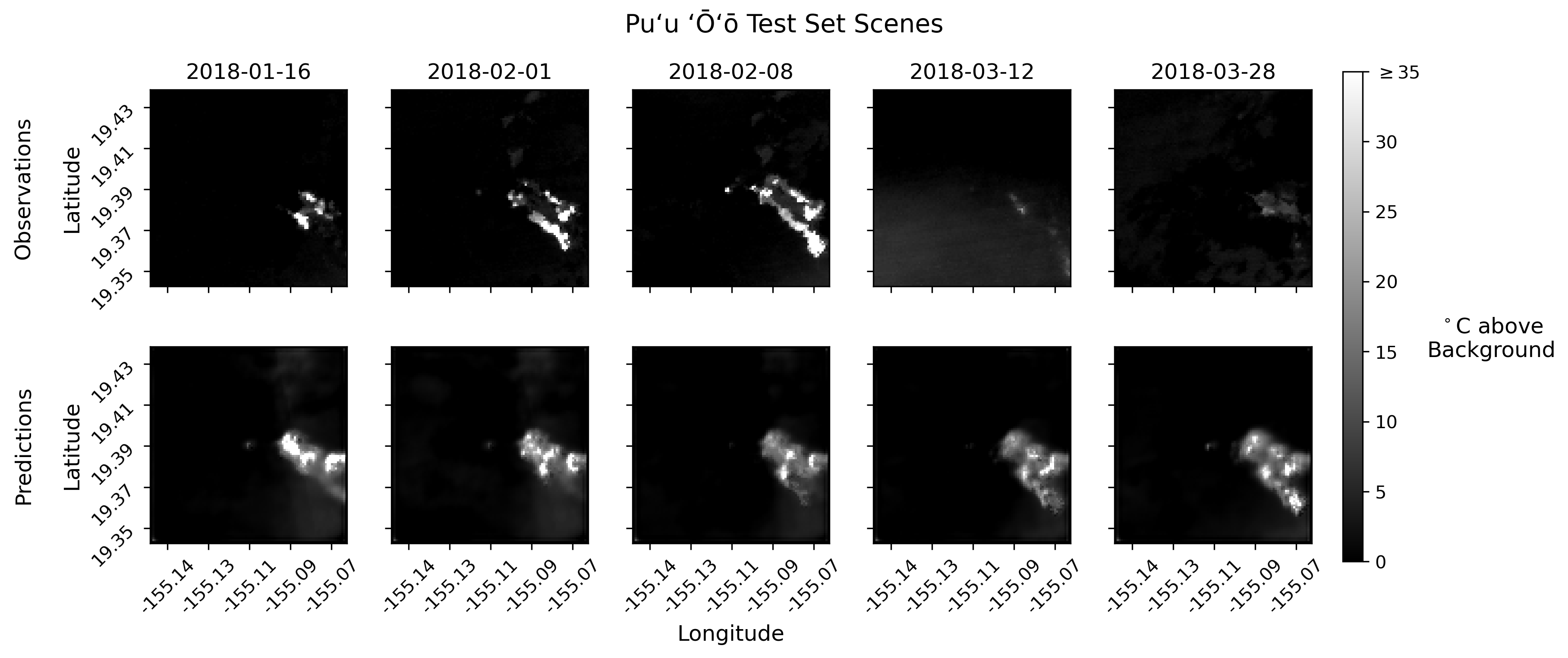}
\caption{Test set image forecasts for Pu'u 'Ō'ō. Notably, the Pu'u 'Ō'ō vent collapsed on April 30, 2018 and there were large lava flows from multiple fissures in the lower East Rift Zone in May, 2018.}
\end{figure*}

\section{Varying the Time between Scenes with the Best Proposed Model}

The existing neural networks used in this paper assume one of three data scenarios: sequential data whose observations are evenly distributed in time, sequential images whose observations are evenly distributed in time, or sequential data whose observations are unevenly distributed in time. The merit of the proposed neural networks is that they represent a combination of the latter two scenarios\textemdash modeling images whose observations are unevenly distributed in time. These proposed models achieved some of the lowest RMSEs compared to existing models and did so efficiently with respect to computation time and number of parameters (see \textbf{Table 1}). Here, we inspect how image forecasts change when the time between scenes is varied for the proposed model that achieved the lowest validation set RMSE on the image forecasting task (ConvLSTM + Time-LSTM + U-net).

We consider six simple modifications to the time differences between scenes to evaluate how these time differences affect the image forecast. First, we consider multiplying and dividing the first time difference in the six-scene input sequence by a factor of $10$; furthermore, we consider the same procedure applied to the last time difference in the six-scene input sequence; and, lastly, we consider adjusting every time difference in this manner. To compare these effects, we consider the mean value of the difference between the originally forecasted scene (with unaltered time differences) and the newly forecasted scene. Additionally, we examine the RMSE of the altered forecast relative to the original forecast. In summary, these two metrics will provide a general direction and magnitude of change.

\begin{table}[t]
\caption{Effect on forecast of changing time differences}
\centering
    \begin{tabular}{c|c|c}
    & Mean of Differenced & RMSE of Altered \\
    Forecast & Forecast & Forecast Relative to \\
    Adjustment & (Adjusted - Original) & Original Forecast \\
    \hline
    First $\Delta$T * 0.1 & 0.002 & 0.003 \\
    Last $\Delta$T * 0.1 & 0.005 & 0.010 \\
    All $\Delta$T * 0.1 & 0.006 & 0.017 \\
    First $\Delta$T * 10 & -0.019 & 0.034 \\
    Last $\Delta$T * 10 & -0.053 & 0.098 \\
    All $\Delta$T * 10 & -0.063 & 0.169 \\
    \end{tabular}
\end{table}

The results presented in \textbf{Table 4} suggest that reducing the time differences by a factor of $10$ has very little effect on image forecasts. The average time difference in the data set is $37$ days, so this likely indicates that the model expects similar changes on the scale of approximately one week to one month. Alternatively, the metrics comparing a change in the forecast are generally an order of magnitude higher when time differences were adjusted in the same manner but by a factor of $10$ (instead of one-tenth). Additionally, predicted pixel values (on average) decreased when time differences were increased. The greatest changes in forecasts were caused by increasing all time differences by a factor of $10$; qualitative inspection of some of the most-changed forecasts showed that there was a tendency for background areas to be forecasted as cooler while volcanically active areas often remained unchanged or were forecasted as hotter. \textbf{Figure 14} displays how some forecasts were different when all of the time differences were increased by a factor of $10$.

\begin{figure}[t]
\centering
\includegraphics[width=\linewidth]{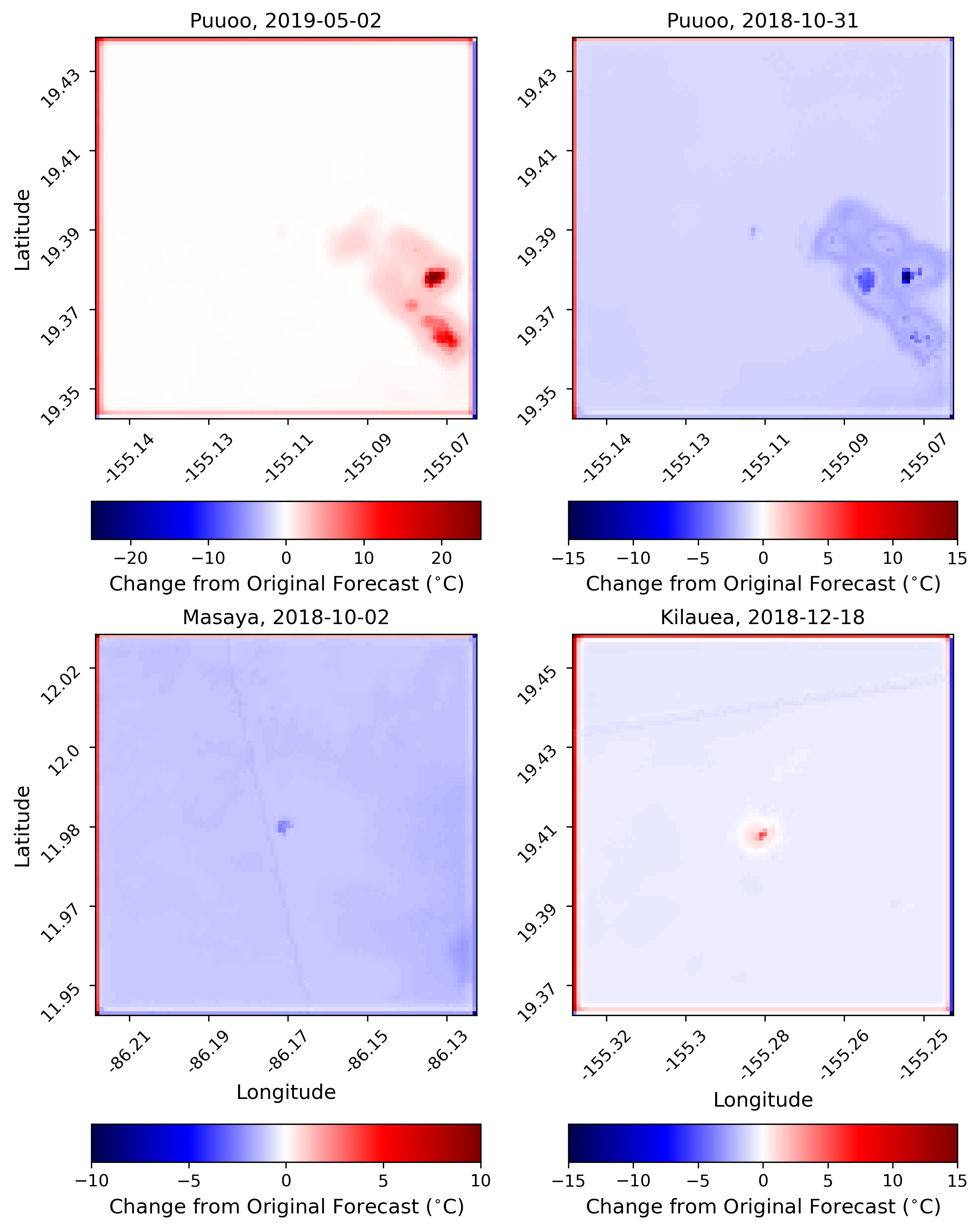}
\caption{Selected examples of how test set forecasts changed when the time between input scenes was increased by a factor of 10. Displayed data are the altered forecasts minus the original forecast after both have been histogram matched to the training set (to account for known issues with the full range of values). Sharp linear features in the bottom images represent mosaicking artifacts.}
\end{figure}

As is common with neural networks, these patterns are difficult to interpret. The tendency to increase average predicted temperatures at short time scales and decrease them at long time scales may be related to the effects of the atmosphere. For example, the thermal signal from clouds does get considered by the forecasts, and sometimes it is passed forward as image features in the forecast. When the time between scenes was increased, the average temperature in the scene decreased, reducing the effects of weather (e.g., relatively warm nighttime clouds) on the long-term forecast. Additionally, one criterion when considering scene viability was whether a volcanic phenomenon was observed through partial cloud cover; for example, large lava flows (particularly at Pu'u 'Ō'ō) were included with partially obstructing conditions. Perhaps the model learned that shorter time between scenes was indicative of more volcanic activity coupled with more nighttime clouds. The more pronounced patterns of forecasting greatly increased or decreased patterns in volcanically active areas are much harder to interpret. It is notable, however, that these patterns are highly spatial, and the existence of differing forecast patterns that emerge when changing the temporal structure is suggestive that the temporal information is valuable and that machine learning algorithms are inclined to find patterns within it.

\section*{Acknowledgment}

This material is based upon work supported by the National Science Foundation Graduate Research Fellowship under Grant No. DGE1255832. Any opinions, findings, and conclusions or recommendations expressed in this material are those of the authors and do not necessarily reflect the views of the National Science Foundation. Additionally, we thank Penn State's Institute of Computational and Data Sciences for both funding and computational resources. Lastly, Helen Greatrex and Manzhu Yu for serving on the advising committee which led to this paper.

\ifCLASSOPTIONcaptionsoff
  \newpage
\fi

\clearpage
\bibliographystyle{IEEEtran}
\bibliography{volcanoes.bib}

\begin{IEEEbiography}[{\includegraphics[width=1in,height=1.25in,clip,keepaspectratio,angle=270]{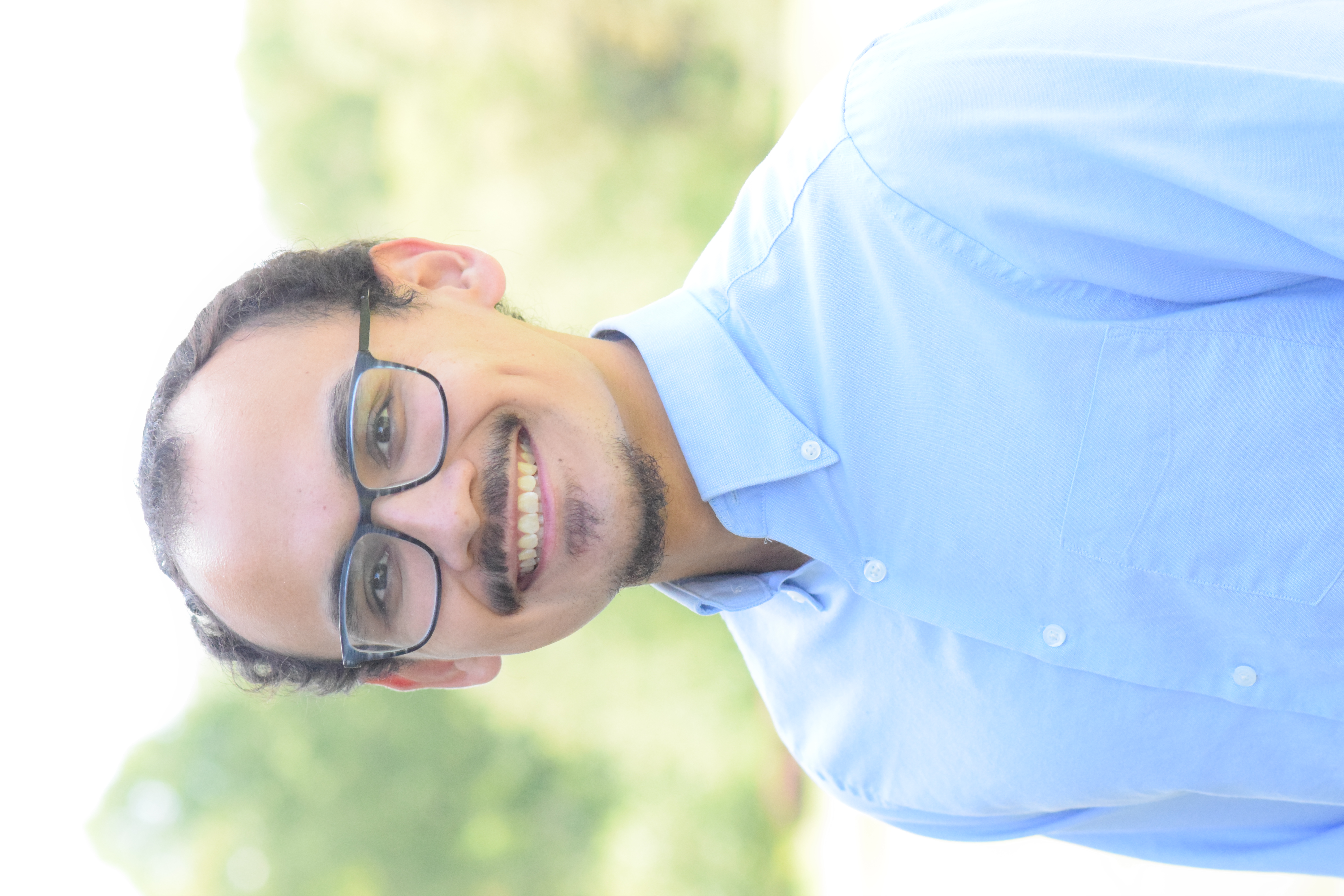}}]{Jeremy Diaz}
received the B.A. degree in ecology and evolutionary biology from the University of Colorado Boulder, Boulder, CO, and the M.S. degree in geography from the Pennsylvania State University, University Park, PA, in 2021.

From 2016 to 2019, he was a student assistant at the Cooperative Institute for Research in Environmental Sciences, Boulder, CO. From 2017 to 2019, he worked with the Earth Lab Analytics Hub applying deep learning to various natural hazards projects. During his time at the Pennsylvania State University (from 2019 to 2021), he was affiliated with the Geoinformatics and Earth Observation Laboratory (GEOlab), the Institute of Computational and Data Sciences, and earned a minor in electrical engineering from the School of Electrical Engineering and Computer Science. His main research interests include applying quantitative methods and computing to natural hazards and earth science research. He is now a Machine Learning Specialist for the United States Geological Survey's Water Mission Area.

Mr. Diaz was a recipient of the 2020 National Science Foundation (NSF) Graduate Research Fellowship (GRF).
\end{IEEEbiography}

\begin{IEEEbiography}[{\includegraphics[width=1in,height=1.25in,clip,keepaspectratio]{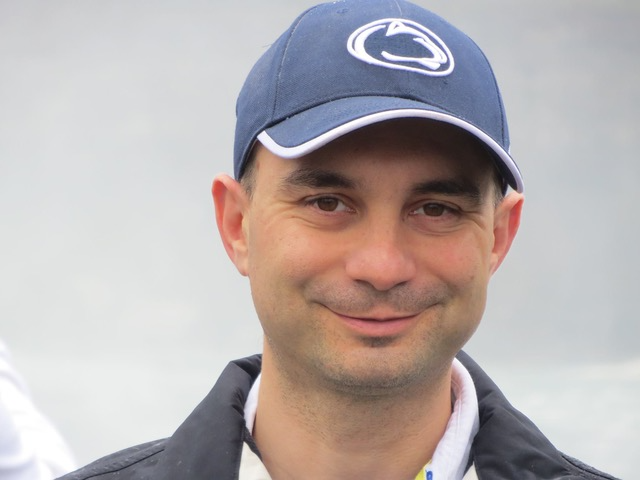}}]{Guido Cervone} received the B.S. degree from the Catholic University of America, Washington, DC, USA, in 1998, and the master’s and Ph.D. degrees in computer science from the George Mason University, Fairfax, VA, USA, in 2000 and 2005, respectively.

He is currently Associate Director of the Institute for Computational and Data Science and Professor in Geography, Meteorology and Atmospheric Science at the Pennsylvania State University, State College, PA, USA. He is also a member of the Earth and Environmental Systems Institute.  His research sits at the intersection of geospatial science, atmospheric science, and computer science. His research focuses on the development and application of computational algorithms for the analysis of remote sensing and numerical modeling data related to environmental hazards and renewable energy.
\end{IEEEbiography}

\begin{IEEEbiography}[{\includegraphics[width=1in,height=1.25in,clip,keepaspectratio]{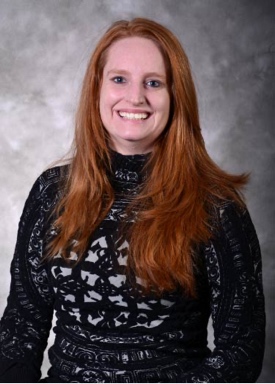}}]{Christelle Wauthier} received a B.S. degree in Engineering, and a M.S. and Ph.D. degrees in geological engineering from the University of Liege, Belgium, in 2003, 2005 and 2011, respectively. She also received a M.S. degree in Volcanology from the University Blaise-Pascal, France, in 2006. She is currently an Associate Professor of Geosciences in the Department of Geosciences and Institute for Computational and Data Sciences, Pennsylvania State University, State College, PA, USA.

Thanks to modeling the surface deformation identified with satellite geodesy, Dr. Wauthier intends solving for the subsurface dynamic characteristics of a volcanic system; i.e., characterize where and how is the magma transported and stored. A thorough study and modeling of ground deformation, together with a stress change analysis, can provide clues on when, where, and how a volcano will erupt, provide insights on magma-tectonic processes and their causality. Her research has deep impacts on society as the geophysical signals registered can be used to assess and mitigate volcanic and seismic hazards.
\end{IEEEbiography}

\end{document}